\documentclass[pdflatex,sn-mathphys-ay]{sn-jnl}

\usepackage{graphicx}%
\usepackage{multirow}%
\usepackage{amsmath,amssymb,amsfonts}%
\usepackage{amsthm}%
\usepackage{mathrsfs}%
\usepackage[title]{appendix}%
\usepackage{xcolor}%
\usepackage{textcomp}%
\usepackage{manyfoot}%
\usepackage{booktabs}%
\usepackage{algorithm}%
\usepackage{algorithmicx}%
\usepackage{algpseudocode}%
\usepackage{listings}%

\usepackage{booktabs}
\usepackage{tabularx}
\usepackage[table]{xcolor}
\usepackage{array}
\usepackage{rotating}

\usepackage{afterpage}

\newcommand{\cls}[1]{\texttt{#1}}
\newcommand{\bench}[1]{\textsc{#1}}

\definecolor{knowncolor}{RGB}{232,244,255}   
\definecolor{crosscolor}{RGB}{255,247,230}   
\definecolor{unseencolor}{RGB}{255,235,238}  

\newcolumntype{K}{>{\columncolor{knowncolor}}c}
\newcolumntype{C}{>{\columncolor{crosscolor}}c}
\newcolumntype{U}{>{\columncolor{unseencolor}}c}

\theoremstyle{thmstyleone}%
\theoremstyle{thmstyletwo}%

\theoremstyle{thmstylethree}%

\begin{document}

\title[Article Title]{COSTA: A Cluster-Centric Paradigm for Annotation-Free Open-Set Semantic Segmentation of Aerial Point Clouds with Domain Shifts}


\author[1,2,3,4]{\fnm{Yanghong} \sur{Lin}}\email{linyanghong21@mails.ucas.ac.cn}
\equalcont{These authors contributed equally to this work.}

\author[2,3,4]{\fnm{Li} \sur{Fang}}\email{lfang@iue.ac.cn}
\equalcont{These authors contributed equally to this work.}

\author[2,3,5]{\fnm{Tianyu} \sur{Li}}\email{2025182050063@whu.edu.cn}

\author[2,3,5]{\fnm{Shudong} \sur{Zhou}}\email{sdzhou@whu.edu.cn}

\author*[2,3]{\fnm{Wei} \sur{Yao}}\email{wyao@iue.ac.cn}

\affil*[1]{\orgdiv{Fujian Institute of Research on the Structure of Matter}, \orgname{Chinese Academy of Sciences}, \orgaddress{\city{Fuzhou}, \postcode{350108}, \state{Fujian}, \country{China}}}

\affil[2]{\orgdiv{Spatial Intelligence and Urban Computing}, \orgname{Institute of Urban Environment, Chinese Academy of Sciences}, \orgaddress{\city{Xiamen}, \postcode{361021}, \state{Fujian}, \country{China}}}

\affil[3]{\orgdiv{State Key Lab for Ecological Security of Regions and Cities}, \orgname{Institute of Urban Environment, Chinese Academy of Sciences}, \orgaddress{\city{Xiamen}, \postcode{361021}, \state{Fujian}, \country{China}}}

\affil[4]{\orgname{University of Chinese Academy of Sciences}, \orgaddress{\city{Beijing}, \postcode{101408}, \state{Beijing}, \country{China}}}

\affil[5]{\orgdiv{School of Resource and Environmental Sciences}, \orgname{Wuhan University}, \orgaddress{\city{Wuhan}, \postcode{430072}, \state{Hubei}, \country{China}}}


\abstract{
Semantic segmentation of aerial point cloud is trapped in a generalization crisis under distinct domain shifts. While test-time adaptation offers a privacy-preserving and computationally efficient way to adapt pre-trained models to unlabeled target-domain data during inference, existing methods, bound to closed-set label assumptions and non-scalable point-wise segmentation pipelines, still struggle with semantic shifts. We ask: can we adapt any given pre-trained aerial point cloud segmentation model to a shifted target domain at the inference phase alone, without additional training, while segmenting target-specific categories beyond the source label space on demand? This paper introduces COSTA, which breaks this limitation by shifting from closed-set point-wise adaptation to cluster-centric open-set semantic propagation. Our core discovery is that, once effectively adapted at test time, the rich feature distribution of aerial point clouds can be distilled into a compact set of well-separated semantic centroids that are transferable across label spaces. COSTA leverages this to reformulate open-set semantic segmentation as a cluster-level propagating process: it first bridges the domain gap through proven test-time adaptation, then groups each batch of target-domain points into a small set of semantic clusters based on the similarity distribution in the adapted feature space, and finally propagates high-confidence pseudo labels obtained from an open-vocabulary vision-language model to all points through cluster-level voting. This cluster-centric paradigm enables test-time adaptation of aerial point clouds under significant domain gaps with mixed semantic shifts. With DALES as the source domain, COSTA enables on-demand segmentation across three aerial point cloud benchmarks with distinct domains and heterogeneous category spaces, achieving up to $70.09\%$ mIoU under this new setting.
}




\maketitle

\section{Introduction}\label{sec1}

Semantic segmentation of aerial point clouds is a fundamental task for urban perception and 3D scene understanding. By assigning semantic labels to individual 3D points, it provides structured scene knowledge for urban 3D modeling~\citep{YAO2011260,  xiang2024efficient, shao2024urban}, aerial-ground collaborative perception~\citep{POLEWSKI2015252,10008011, hou2026agc} and autonomous UAV navigation~\citep{wang2025towards,gao2026openfly}. 

In recent years, deep neural networks have emerged as leading methods for this task by learning high-level representations from irregular point clouds~\citep{hu2022sensaturban,li2025deep}. Despite remarkable advances, existing methods are still largely built upon a closed-set and domain-specific learning paradigm~\citep{choy20194d,kolodiazhnyi2024oneformer3d,wu2024point}. Most models are trained and evaluated on specific 3D benchmarks with predefined semantic categories and densely annotated point-wise labels, often ignoring generalization and transferability across different 3D domains. However, various 3D domains may differ drastically in data distributions, spatial scales, point densities, and noise patterns, leading to significant degradation of domain-specific pre-trained models~\citep{wang2025test,zou2026good}. To address this issue, domain adaptation (DA) has been widely explored to alleviate such domain shifts by transferring knowledge from a labeled source domain to an unlabeled or weakly labeled target domain. Among different adaptation settings, test-time adaptation (TTA) adapts a pre-trained model directly to unlabeled target-domain data during inference, without accessing the source-domain data or launching an additional training process. Due to its efficiency, low computational cost, and ability to avoid the potential security risks associated with the source, TTA has attracted increasing attention and has been explored for aerial point clouds in recent years~\citep{wang2025test,gao2026source}. While offering a promising solution to this problem, existing TTA methods remain confined to closed-set adaptation, due to closed-set label assumptions and non-scalable point-wise segmentation frameworks. In real‑world deployments, shifts in geographic location, data distribution, and user requirements naturally introduce target data that contain not only known classes overlapping with the source, but also novel classes with finer granularity or previously unseen semantics. This open‑set challenge fundamentally limits the applicability of current TTA techniques, whose closed‑set predictions inherently fail to accommodate emerging classes. Although recent advances~\citep{li2023robustness,gao2024unified,zou2026good} have begun to acknowledge the open-set challenge, they typically treat unseen classes merely as out-of-distribution (OOD) samples, with the aim of preventing degradation of adaptation in known classes. As a result, segmentation remains restricted to categories that overlap with the source. 

Open-vocabulary vision-language models (VLMs) provide a promising opportunity to break this semantic bottleneck. By conditioning segmentation on natural language prompts, VLMs can segment or localize categories beyond the source, enabling on-demand semantic understanding. Nevertheless, directly transferring open-vocabulary image segmentation to aerial point clouds is non-trivial. Most powerful VLMs are trained and operate primarily on 2D natural images~\citep{carion2025sam3segmentconcepts,yuan2025sa2va}, while aerial point clouds are sparse, irregular, and three-dimensional, inevitably resulting in a substantial modality gap. A straightforward and widely adopted strategy is to exploit images aligned well with point clouds~\citep{peng2023openscene,yang2024regionplc,jiang2024open}. These images are fed into 2D VLMs to obtain category-specific 2D prediction masks or visual features, which are then back-projected onto the corresponding 3D points as pseudo labels or feature-alignment supervision for 3D semantic segmentation. However, these attempts to circumvent the modality gap are fundamentally limited in aerial scenarios. Drone or satellite images inevitably differ from natural images in viewpoint, spatial resolution, and imaging geometry, and are subject to occlusion and incomplete visibility~\citep{wang2026openurban3d,lin2026towards}. As a result, the 2D supervision derived from VLMs is often incomplete, noisy, and spatially inconsistent. Direct domain adaptation on the  VLMs to mitigate these discrepancies is also a suboptimal solution: it requires updating a much larger parameter space, introduces substantial computational overhead, and may not guarantee direct improvements in 3D point cloud segmentation. Moreover, in most aerial point cloud benchmarks, the frequent absence of well-aligned synchronously acquired images further limits the direct applicability of 2D VLMs~\citep{xu2025cityseg,wang2026openurban3d}. Even for the few benchmarks equipped with auxiliary imagery, such as Hessigheim 3D~\citep{kolle2021hessigheim}, the large temporal gap between image and point cloud acquisition makes the supervision for dynamic or time-sensitive categories, such as vehicles and growing vegetation, unreliable or invalid.

\begin{figure*}[h]
\centering
\includegraphics[width=0.9\textwidth]{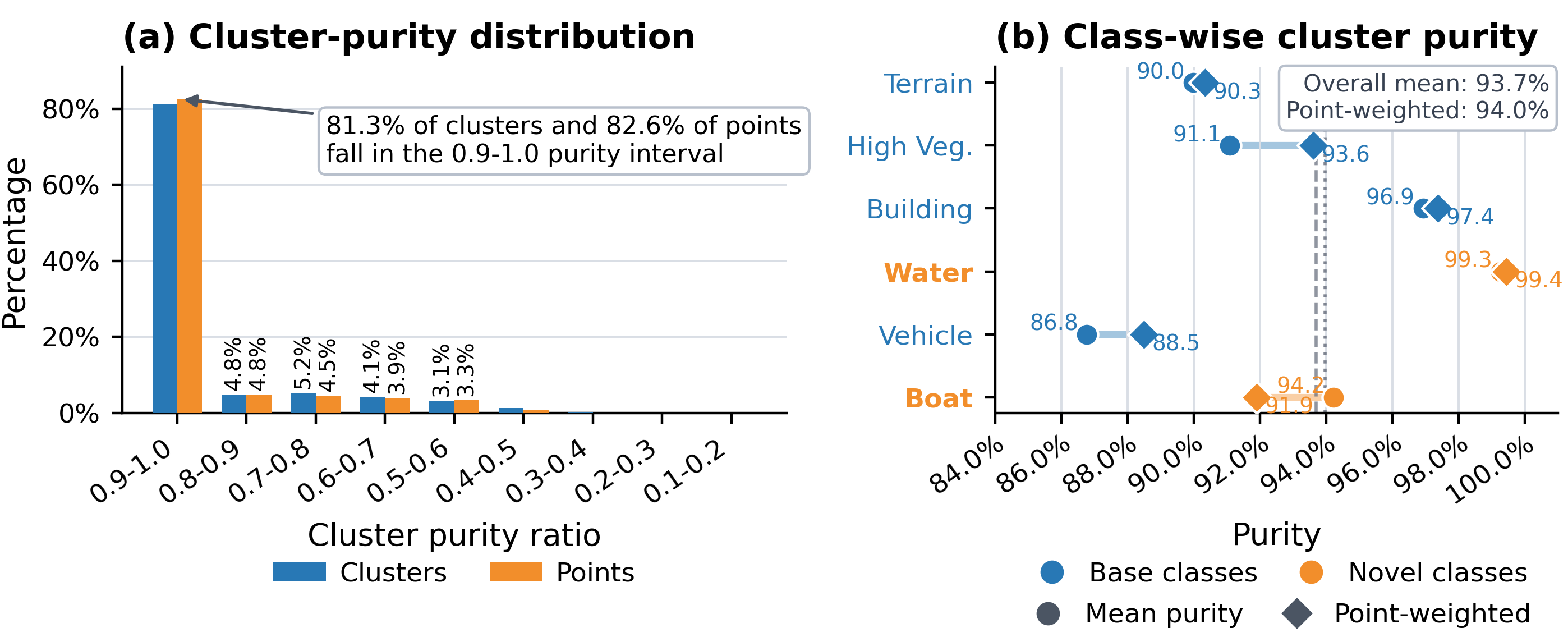}
\caption{With DALES as the source domain and SUM as the target domain, we partition the target-domain data with K-Means in TTA feature space. The results show that most clusters exhibit high semantic purity: over 80\% of the clusters have a proportion of same-class points ranging from 0.9 to 1.0. Notably, this observation extends beyond the source label space.}\label{fig:cluster_purity_intro}
\end{figure*}

To this end, we consider a more realistic setting, i.e. annotation-free open-set test-time adaptation (OSTTA): during inference, a pre-trained aerial point cloud segmentation model is required to adapt to an unlabeled target domain without source data or additional training, while simultaneously recognizing target-specific categories beyond the source label space. To the best of our knowledge, it has not been explored before. We posit that the key lies in a paradigm shift from non-scalable closed-set point-wise adaptation and prediction to cluster-centric open-set semantic voting. Our core observation is that, after effective test-time adaptation, points with similar semantics in the target domain tend to form compact local clusters in the adapted feature space, and each cluster exhibits high semantic purity (Fig. \ref{fig:cluster_purity_intro}). This observation extends beyond the source-domain label space and helps counteract the noisy pseudo-labels produced by VLMs. Inspired by this, we introduce COSTA, a novel cluster-centric paradigm for annotation-free open-set semantic segmentation of aerial point clouds under significant domain gaps with mixed semantic shifts. 

COSTA exploits the complementary strengths of the TTA 3D segmentation network and open-vocabulary VLMs using clusters as a semantic bridge. The 3D network provides a target-adapted feature space where semantically coherent clusters can be discovered beyond the source label space, while the VLMs supply flexible category-level evidence on demand. Rather than blindly trusting noisy point-wise VLMs predictions or being restricted to a closed-set classifier head, COSTA aggregates reliable open-vocabulary cues at the cluster level and propagates them back to points. This design makes open-set semantic adaptation both label-extensible and robust to noisy pseudo labels.

Our contributions are summarized as follows.

\begin{itemize}
    \item We introduce COSTA, the first cluster-centric open-set test-time adaptation framework for aerial point cloud semantic segmentation. By integrating a lightweight test-time adaptation strategy that optimizes learnable BN parameters via entropy minimization, information maximization, and OOD discovery, COSTA enables a source-pretrained model to adapt to unlabeled target domains during inference and segment target-specific categories beyond the source label space on demand.
    \item We design an adaptive hybrid-view projection module that combines oblique virtual-camera and BEV projections, allowing seamless application of 2D vision-language models to 3D aerial point clouds.
    \item We propose a reliability-aware cluster-centric propagation mechanism that bridges the adapted 3D feature space with open-vocabulary VLMs evidence. By aggregating multi-view cues through source-specific response thresholds and point-level reliability estimation, it propagates high-confidence pseudo labels via cluster-level voting, enabling robust generalization under domain gaps with mixed semantic shifts.
    \item We establish three heterogeneous aerial point cloud benchmarks covering two representative semantic shifts: unseen classes and cross-granularity classes. Extensive experiments demonstrate that COSTA enables annotation-free on-demand open-set test-time adaptation segmentation and achieves SOTA performance under this setting.
\end{itemize}

\section{Related Work}\label{sec2}
\subsection{Point Cloud Semantic Segmentation}

The interpretation of 3D point clouds has evolved from basic geometric grouping to complex semantic scene parsing. Recently, point-based networks have become a dominant paradigm for directly learning from unordered point sets. PointNet~\citep{qi2017pointnet} first introduced symmetric aggregation functions to achieve permutation invariance, while PointNet++~\citep{qi2017pointnet++} further incorporated hierarchical local feature learning to better capture fine-grained geometric structures. Later works improved point-wise representation learning through more effective local aggregation and scalable sampling strategies. For example, RandLA-Net~\citep{hu2020randla} employs random sampling with local feature aggregation to efficiently process large-scale point clouds, and PointNeXt~\citep{qian2022pointnext} revisits PointNet++ with stronger training and scaling strategies, showing that carefully designed optimization recipes and scalable architectures can substantially improve segmentation performance.

Another line of research converts irregular point clouds into structured or semi-structured representations to enable convolutional processing. Sparse convolutional networks, such as MinkowskiNet~\citep{choy20194d}, operate on sparse tensors and significantly reduce computation by applying convolutions only to occupied locations. KPConv~\citep{thomas2019kpconv} defines convolution kernels directly in Euclidean space through learnable kernel points, enabling flexible local geometric modeling without voxelizing the input. Related sparse point-voxel designs, such as SPVNAS/SPVCNN~\citep{tang2020searching}, further balance point-level geometric precision and voxel-level computational efficiency. These convolution-based methods have achieved strong performance on both indoor and outdoor benchmarks and remain widely used as backbones for large-scale 3D point cloud segmentation.

More recently, Transformer-based architectures have advanced fully supervised point cloud segmentation by modeling long-range dependencies and global context. Point Transformer and its successors~\citep{zhao2021point,wu2022point,wu2024point} adapt self-attention to irregular 3D data and achieve strong performance through local attention, grouped vector attention, and scalable point serialization. Swin3D~\citep{yang2025swin3d} introduces sparse-window attention and pretraining strategies for 3D scene understanding, while OctFormer~\citep{wang2023octformer} leverages octree-based attention to improve scalability on large point clouds. Recent unified frameworks such as OneFormer3D~\citep{kolodiazhnyi2024oneformer3d} further extend Transformer-based designs to semantic, instance, and panoptic segmentation within a single architecture. These methods demonstrate the strong representation capacity of the transformer-based backbone.

Despite these advances, point cloud segmentation remains constrained by a closed-set and annotation-intensive learning paradigm. The model is trained on a specific benchmark to predict a fixed set of predefined classes, and its performance largely depends on the availability of dense point-wise annotations in the target domain~\citep{wang2022new}. When deployed in aerial or urban scenes with domain shifts (e.g., different sensors, regions, classes), supervised models typically suffer significant performance degradation or fail entirely, necessitating costly re-annotation and retraining. This limitation motivates our open-set test-time adaptation setting, where a source-pretrained model is adapted to unlabeled target point clouds during inference while supporting user-specified classes beyond the source.

\subsection{Test-time Adaptation}

Among various approaches for addressing covariate shifts induced by domain gaps, test-time adaptation has attracted increasing attention due to its challenging setting, where a source-trained model is adapted to the target distribution during inference accessing only unlabeled target data. Early work focused primarily on 2D vision, especially for image classification and semantic segmentation. Representative methods such as TENT~\citep{wang2021tent} minimize prediction entropy by updating normalization statistics and affine parameters of the BN layer during inference, while EATA~\citep{niu2022efficient} improves efficiency and stability through reliable sample selection and anti-forgetting regularization. Continual TTA methods further consider non-stationary target streams, where CoTTA~\citep{wang2022continual} reduces error accumulation by using weight-averaged predictions and stochastic restoration. More recently, TTA has been extended to 3D point cloud understanding. GIPSO~\citep{saltori2022gipso} is the first method for TTA 3D point cloud segmentation by introducing geometrically informed pseudo-label propagation for online adaptation. HGL~\citep{zou2024hgl} exploits local-global hierarchical pseudo-label strategy to improve TTA 3D segmentation. For geospatial point clouds, recent work adapts pretrained models by updating the statistics and learnable affine parameters of BN layer with self-supervised learning~\citep{wang2025test}. TTT-KD~\citep{weijler2025ttt} further explores test-time training for 3D semantic segmentation by distilling knowledge from 2D foundation models. While promising, most existing TTA methods are designed under a closed-set assumption, where the source and target domains share the same semantic. These methods mainly address covariate shifts caused by sensor changes, weather, or density variations, but do not explicitly handle target classes that are absent from or inconsistent with the source.

Recently, some works have paid more attention to improving the robustness of TTA methods. NOTE~\citep{gong2022note} uses instance-aware batch normalization and prediction-balanced reservoir sampling to alleviate biased online updates, while RoTTA~\citep{yuan2023robust} introduces robust teacher-student adaptation with time-aware reweighting for dynamic scenarios. SAR~\citep{niu2023towards} analyzes several practical failure cases of TTA, including mixed distribution shifts, small batch sizes, and imbalanced online label distributions, and stabilizes the minimization of entropy by filtering unreliable samples and encouraging flat minima. Another line of work studies open-world TTA, where the test data contains unknown classes. OWTTT~\citep{li2023robustness} improves robustness by pruning strong OOD samples and expanding prototypes. UniEnt~\citep{gao2024unified} performs entropy minimization on in-distribution (ID) samples and entropy maximization on pseudo-covariate-shifted OOD samples. Multimodal and stabilized open world TTA methods further refine unknown-sample filtering and entropy-aware optimization~\citep{dong2025towards,lee2025stabilizing}. Recently, GOOD~\citep{zou2026good} introduced the first open world TTA method for 3D point cloud semantic segmentation, which enhances the discrimination between ID and OOD regions by integrating a novel confidence metric and prototype representation. However, these robust TTA methods focus on improving the model’s ability to distinguish between ID and OOD samples and preventing OOD or unknown samples from degrading the adaptation of known classes. This fundamentally differs from our OSTTA setup, which aims not only to preserve the model’s adaptability to known classes but also to recognize user-specified target classes beyond the source.

\subsection{Open-Vocabulary Semantic Segmentation}
To overcome closed-set constraints, open-vocabulary segmentation has been developed using large-scale vision-language models (VLMs). Early works focused on 2D vision, primarily using 2D VLMs pre-trained on image-text pairs to perform open-vocabulary classification or segmentation of images. OpenSeg~\citep{10.1007/978-3-031-20059-5_31} and MaskCLIP~\citep{10205470} exploit CLIP-like representations for open-vocabulary semantic, instance, or panoptic segmentation. Subsequent methods further improve the spatial precision and semantic alignment of open-vocabulary segmentation. ODISE~\citep{xu2023open} leverages text-to-image diffusion representations for open-vocabulary panoptic segmentation, while CAT-Seg~\citep{cho2024cat} formulates segmentation as a cost aggregation between dense image features and textual embeddings. These methods demonstrate the strong potential of language-conditioned visual models to generalize arbitrary classes. 

Driven by the success of 2D VLMs, recent efforts have explored how to transfer open-vocabulary semantics from 2D foundation models to 3D representations. A common strategy is distilling knowledge from 2D VLMs into 3D backbones or aligning class-agnostic 3D proposals with multi-view image features. OpenScene~\citep{peng2023openscene} aligns 3D point features with CLIP image and text embeddings through multi-view distillation. PLA~\citep{ding2023pla} introduces caption-assisted language supervision to associate 3D representations with semantically richer textual descriptions. SEAL~\citep{liu2023segment} and OV3D~\citep{jiang2024open} further exploit 2D--3D correspondences and foundation models to improve open-vocabulary 3D semantic segmentation. Region-level methods, such as RegionPLC~\citep{yang2024regionplc} and OpenMask3D~\citep{takmaz2023openmask3d}, aggregate multi-view language-aligned features over class-agnostic 3D regions or instance masks, thereby improving robustness to noisy point-wise alignment. However, these methods rely on spatially and temporally well-aligned RGB images, which are often not satisfied in aerial point cloud benchmarks. OpenUrban3D~\citep{wang2026openurban3d} leverages multi-view rendered point clouds and vision–language feature distillation to achieve open vocabulary semantic understanding in large-scale urban scenes. More recently, some works, such as CitySeg~\citep{xu2025cityseg}, have attempted to replicate the success of 2D VLMs by directly incorporating textual semantics during large-scale city-scale point cloud training. However, due to the scarcity of 3D-text datasets, 3D point cloud open-vocabulary semantic segmentation models typically have fewer parameters and weaker generalization performance compared to 2D VLMs. The more pronounced disparities in perspective, geographical location, and density in aerial point clouds further amplify this issue~\citep{wang2025test}. Moreover, the costly retraining inevitably limits the practicality of such methods.

\begin{figure}[h]
\centering
\includegraphics[width=0.96\textwidth]{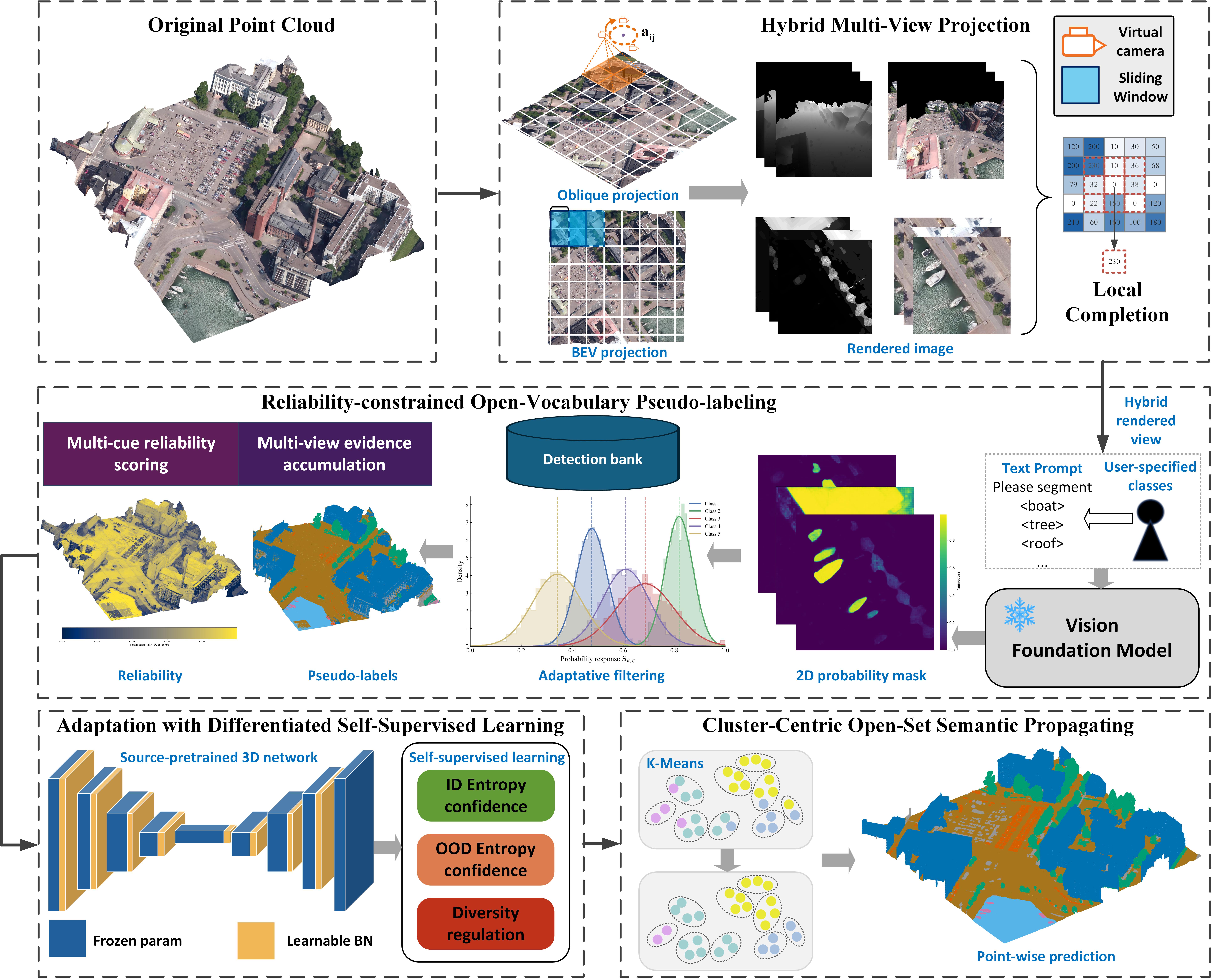}
\caption{Overview of the proposed COSTA for OSTTA. We first perform hybrid multi-view rendering on target point cloud by exploiting complementary BEV and oblique virtual camera projection. This yields hybrid view set with pixel-to-point correspondences for seamless VLMs semantic querying without well-aligned images. Given user-specified target classes, we prompt VLMs to generate open-vocabulary 2D probability masks, which are filtered adaptively and fused through reliability-constrained multi-view evidence accumulation. This provides reliability-aware pseudo labels for the next step. Given the pseudo labels, we adapt source-pretrained 3D model to the target data through differentiated self-supervised learning during inference. Finally, we perform reliability-weighted cluster-level semantic propagation by clustering the adapted point features. The framework produces on-demand open-set semantic segmentation on the target domain without source-data replay, target annotations, or offline retraining.}\label{fig:overview}
\end{figure}

\section{Methods}\label{sec3}

\subsection{Overview}
Given an aerial point cloud segmentation model $M_s$ pre-trained on a source domain $D_s$ with source classes $\mathcal{C}_s$, our objective is to design an open-set test-time adaptation method $f$ that adapts $M_s$ to an unlabeled target domain $D_t$ and predicts point-wise semantic labels during inference. Unlike conventional closed-set TTA, the user can specify a set of target classes $\mathcal{C}_t$ on demand via text queries, where $\mathcal{C}_t$ is not restricted to the source classes (i.e., $\mathcal{C}_t$ may contain categories beyond $\mathcal{C}_s$). Formally, the method $f$ produces an adapted model $M_t$ and the corresponding point-wise predictions $\hat{Y}_t$ as
\begin{equation}
M_t,\;\hat{Y}_t = f(M_s,\,D_t,\,\mathcal{C}_t).\label{eq1}
\end{equation}
with the objective of minimizing the expected risk over the target distribution while producing predictions under semantic shifts.

To address the challenge of cross-domain generalization under semantic shifts, we propose a novel cluster-centric framework for annotation-free open-set test-time adaptation, namely COSTA (see Fig.~\ref{fig:overview}). First, raw point clouds are rendered into a complementary 2D image via adaptive hybrid multi-view projection, including oblique virtual camera projections and bird’s-eye-view (BEV) projections, thereby avoiding dependency on well-aligned synchronously acquired images. By querying with user-specified text prompts, the VLMs generate probabilistic 2D masks from the rendered image for arbitrary classes demanded in the target domain. To recover high-quality 3D pseudo-supervision, we further introduce a reliability-aware pseudo-labeling strategy based on class-wise adaptive detection banks and multi-view evidence accumulation strategy, which back-projects and fuses VLMs-derived probability masks into point-wise pseudo labels. Subsequently, we directly optimize the batch normaolization (BN) layers of the source pre-trained model through a differentiated self-supervised entropy optimization strategy on pseudo-ID and pseudo-OOD samples from the test data, which aims to learn target domain-specific knowledge. Finally, the adapted point-wise features are grouped by K-Means into semantically coherent clusters, followed by open-set reliability-weighted semantic voting over all trustworthy bank points within each cluster.
    
\subsection{Pseudo-labeling with Reliability Constraint}
A direct approach to utilizing VLMs in open-set TTA is to fetch predictions for target classes from synchronously acquired, well-calibrated images and then establish the correspondence between the point cloud and the image with prior calibration information, thereby inferring the point predictions from the VLMs. However, most aerial point cloud datasets either lack such calibrated imagery or exhibit significant temporal misalignment, severely limiting the applicability of VLMs. Moreover, these raw predictions are noisy and can introduce error accumulation from the 2D VLMs into the 3D models. Therefore, we introduce an effective, reliability-aware pseudo-labeling strategy for filtering VLMs predictions inferred from hybrid rendered images.

\subsubsection{Hybrid Multi-View Rendering}\label{sec:hybrid}
To address the challenge of the frequent absence of corresponding images in aerial point clouds, we design an adaptive hybrid multi-view projection module that synthesizes information-rich rendered images. The hybrid multi-view projection module constructs a hybrid view set that fully exploits the respective advantages of tilted oblique virtual-camera and BEV projections (see Algorithm~\ref{alg:hybrid_projection}). The former captures comprehensive side-view geometric appearance of vertical structures from diverse perspectives, while the latter provides stable top-down observations that are particularly effective for small top-visible objects. Unlike fixed projection parameters, the projection configuration is derived from the spatial extent of each scene, making the rendered images adaptive to scene scale and point density. 

Formally, given a target point cloud $\mathcal{P}_t=\{p_i\}_{i=1}^{N}$, where $p_i=[x_i,y_i,z_i,r_i,g_i,b_i]$, we compute its axis-aligned bounds and scene scale as
\begin{equation}
W=x_{\max}-x_{\min},\quad
L=y_{\max}-y_{\min},\quad
H=z_{\max}-z_{\min},\quad
s=\sqrt{WL}.
\label{eq:scene_scale}
\end{equation}

\textbf{Tiled oblique virtual-camera projection:} For the tilted virtual-camera projection, we partition the horizontal plane into $K_l\times K_l$ local regions. For each region $(i,j)$ with $i,j\in\{1,\dots,K_l\}$, a camera anchor $\mathbf{a}_{ij}$ and a target point $\mathbf{t}_{ij}$ are defined adaptively based on the scene scale to guide the viewpoint of the virtual camera as 
\begin{equation}
\mathbf{a}_{ij}=
\left(
x_{\min}+\frac{iW}{K_l+1},
y_{\min}+\frac{jL}{K_l+1},
z_{\min}+H+\frac{s}{2K_l}
\right),
\label{eq:camera_anchor}
\end{equation}
\begin{equation}
\mathbf{t}_{ij}=
\left(
a_{ij}^{x},
a_{ij}^{y},
z_{\min}+H+\frac{s}{\rho K_l}
\right),
\label{eq:camera_target}
\end{equation}
Multiple virtual cameras orbit around the camera anchor $\mathbf{a}_{ij}$ along a circular horizontal path of radius $R_l$ and are oriented toward the target point $\mathbf{t}_{ij}$. Thus, the world coordinates of the virtual camera orbiting camera anchor $\mathbf{a}_{ij}$ are computed as
\begin{equation}
\mathbf{c}_{ijk}
=
\mathbf{a}_{ij}
+
\left(
R_l \cos\theta_k,R_l\sin\theta_k,0
\right),
\quad
R_l=\frac{s}{\rho K_l}.
\label{eq:camera_position}
\end{equation}
where the hyper-parameter $\rho$ controls the radius of the camera's circular orbit. To ensure that each virtual camera covers the entire local region, the field of view (FOV) of each virtual camera is set adaptively. Let $d_l=\sqrt{(W/K_l)^2+(L/K_l)^2}$ be the diagonal length of a local region cell and $h_l=a_{ij}^{z}-t_{ij}^{z}$ be the vertical camera-target distance. The FOV is given by
\begin{equation}
\mathrm{FOV}_{ij}
=
\min\left(
2\arctan\frac{d_l}{2h_l},\;120^\circ
\right).
\label{eq:fov}
\end{equation}
which ensures that each local view contains sufficient spatial context while preventing extreme perspective distortion with safety margin.

Given the camera position, target point, and FOV, each virtual camera is represented by a standard pinhole camera model. The model consists of an extrinsic matrix, which transforms points from the world coordinate system to the camera coordinate system, and an intrinsic matrix, which maps camera-coordinate of points onto the image plane.

The extrinsic matrix is first determined by the camera pose. Specifically, for a virtual camera located at $\mathbf{c}$ and looking at target point $\mathbf{t}$, we construct an orthonormal camera coordinate system using the world-up direction $\mathbf{u}=[0,0,1]^\top$. The forward, right, and image-downward axes are computed as
\begin{equation}
\mathbf{f}=\frac{\mathbf{t}-\mathbf{c}}{\|\mathbf{t}-\mathbf{c}\|_2},\quad
\mathbf{r}=\frac{\mathbf{f}\times\mathbf{u}}{\|\mathbf{f}\times\mathbf{u}\|_2},\quad
\mathbf{d}=\frac{\mathbf{f}\times\mathbf{r}}{\|\mathbf{f}\times\mathbf{r}\|_2}.
\label{eq:camera_axes}
\end{equation}
Here, $\mathbf{f}$ represents the viewing direction, while $\mathbf{r}$ and $\mathbf{d}$ define the horizontal and vertical axes of the image plane, respectively. The world-to-camera rotation and translation are then obtained as
\begin{equation}
\mathbf{R}=
\begin{bmatrix}
\mathbf{r}^{\top}\\
\mathbf{d}^{\top}\\
\mathbf{f}^{\top}
\end{bmatrix},
\quad
\mathbf{t}_{\mathrm{cam}}=-\mathbf{R}\mathbf{c},
\label{eq:extrinsic_rt}
\end{equation}
forming the extrinsic matrix
\begin{equation}
\mathbf{E}=
\begin{bmatrix}
\mathbf{R} & \mathbf{t}_{\mathrm{cam}}
\end{bmatrix}.
\label{eq:extrinsic}
\end{equation}

The intrinsic matrix is determined by the image resolution and the FOV. For an image of size $W_I\times H_I$, the focal length is $f_I=(W_I/2)/\tan(\mathrm{FOV}/2)$, yielding the intrinsic matrix
\begin{equation}
\mathbf{K}=
\begin{bmatrix}
f_I&0&W_I/2\\
0&f_I&H_I/2\\
0&0&1
\end{bmatrix}.
\label{eq:intrinsic}
\end{equation}
A point is projected onto the image plane through this camera model. During rendering, we maintain a depth buffer: when multiple points fall into the same pixel, only the point with the nearest camera depth is preserved. This produces a rendered RGB image, a depth map, and a visibility-aware point-to-pixel correspondence.

\textbf{BEV projection:} For BEV projection, we set a sliding window to process overlapping point cloud tiles. Given $s_g$, $G$, and  $\Delta_g$ that controls the scale, window size, and moving steps of the sliding windows, we extract points in the current sliding windows to reduce the following data processing volume. For the $m$-th BEV tile with origin $\mathbf{o}_m=(o_m^x,o_m^y)$, a point $p_i$ is rasterized into pixel coordinates $(u_{i,m}^{\mathrm{bev}}, v_{i,m}^{\mathrm{bev}})$ for BEV RGB and depth queries, where $u$ and $v$ are column and row indices, respectively:
\begin{equation}
u_{i,m}^{\mathrm{bev}}
=
\left\lfloor\frac{x_i-o_m^x}{s_g}\right\rfloor,\quad
v_{i,m}^{\mathrm{bev}}
=
\left\lfloor\frac{y_i-o_m^y}{s_g}\right\rfloor.
\label{eq:bev_raster}
\end{equation}
Since each tile covers an area of $G\times G$ meters and one pixel corresponds to $s_g$ meters, the resulting BEV image has a spatial resolution of $\lfloor G/s_g \rfloor \times \lfloor G/s_g \rfloor$ pixels. Only points whose computed pixel indices fall within this range are considered for the tile. If multiple points fall into the same BEV pixel, the topmost point is retained. The retained point defines both the BEV RGB value and the depth map indexed by $(u_{i,m}^{\mathrm{bev}}, v_{i,m}^{\mathrm{bev}})$. This top-down projection is particularly suitable for aerial scenarios, which reduces the occlusion effects caused by oblique projections on low or small objects.

To mitigate sparse holes caused by low point density, we further apply pixel-level completion for rendered images, especially for the internal areas and margins. For BEV images, missing pixels are filled by iterative $3\times3$ 2D max-pooling:
\begin{equation}
X^{(l+1)}(u,v)=
\begin{cases}
X^{(l)}(u,v), & X^{(l)}(u,v)\neq\varnothing,\\
\max\limits_{(a,b)\in\mathcal{N}_3(u,v)}X^{(l)}(a,b), & \mathrm{otherwise}.
\end{cases}
\label{eq:bev_completion}
\end{equation}
Here, $X^{(l)}$ denotes the BEV map after the $l$-th completion iteration, where $l=0$ corresponds to the initially rasterized sparse map. For virtual-camera views, completion is similar and depth-guided. At each iteration, for a missing pixel $(u,v)$, COSTA searches its $3\times3$ neighborhood and selects the valid neighboring pixel with the smallest positive depth. The missing RGB value is then duplicated from this nearest visible neighbor, which reduces background bleed-in and preserves object boundaries. The completion only improves visual continuity for VLMs inference; the subsequent back-projection still relies on the original visibility-aware correspondences.

\begin{algorithm}[!htp]
\caption{Hybrid Multi-View Projection}
\label{alg:hybrid_projection}
\begin{algorithmic}[1]
\Require Target point cloud $\mathcal{P}_t=\{p_i=[x_i,y_i,z_i,r_i,g_i,b_i]\}_{i=1}^{N}$; local grid number $K_l$; camera angles $\Theta=\{\theta_k\}_{k=1}^{K_\theta}$; orbit factor $\rho$; image size $(W_I,H_I)$; BEV scale $s_g$; BEV tile size $G$; BEV stride $\Delta_g$; completion iterations $L_c$
\Ensure Hybrid rendered view set $\mathcal{V}=\mathcal{V}_{\mathrm{obl}}\cup\mathcal{V}_{\mathrm{bev}}$

\Statex
\State \textbf{Step 1: Compute adaptive scene scale}
\State Compute scene bounds $(x_{\min},x_{\max},y_{\min},y_{\max},z_{\min},z_{\max})$ from $\mathcal{P}_t$
\State $W\gets x_{\max}-x_{\min}$, $L\gets y_{\max}-y_{\min}$, $H\gets z_{\max}-z_{\min}$, $s\gets\sqrt{WL}$
\State Initialize $\mathcal{V}_{\mathrm{obl}}\gets\emptyset$ and $\mathcal{V}_{\mathrm{bev}}\gets\emptyset$

\Statex
\State \textbf{Step 2: Generate tiled oblique virtual-camera views}
\State $d_l\gets \sqrt{(W/K_l)^2+(L/K_l)^2}$
\For{$i=1$ to $K_l$}
    \For{$j=1$ to $K_l$}
        \State $\mathbf{a}_{ij}\gets \left(x_{\min}+\frac{iW}{K_l+1},\,y_{\min}+\frac{jL}{K_l+1},\,z_{\min}+H+\frac{s}{2K_l}\right)$ \Comment{camera anchor}
        \State $\mathbf{t}_{ij}\gets \left(a_{ij}^{x},a_{ij}^{y},z_{\min}+H+\frac{s}{\rho K_l}\right)$ \Comment{target point}
        \State $R_l\gets \frac{s}{\rho K_l}$, $h_l\gets a_{ij}^{z}-t_{ij}^{z}$
        \State $\mathrm{FOV}_{ij}\gets \min\left(2\arctan\frac{d_l}{2h_l},120^\circ\right)$ \Comment{field of view}
        \ForAll{$\theta_k\in\Theta$}
            \State $\mathbf{c}_{ijk}\gets \mathbf{a}_{ij}+\left(R_l\cos\theta_k,R_l\sin\theta_k,0\right)$ \Comment{coord of virtual camera}
            \State $\mathbf{E}_{ijk}\gets \mathrm{LookAt}(\mathbf{c}_{ijk},\mathbf{t}_{ij})$ \Comment{extrinsic of virtual camera}
            \State $\mathbf{K}_{ijk}\gets \mathrm{Intrinsic}(W_I,H_I,\mathrm{FOV}_{ij})$ \Comment{intrinsic of virtual camera}
            \State Render RGB image $I$, depth map $D$, and correspondence map $\Gamma$ with depth buffering
            \State $(I,D)\gets \mathrm{DepthGuidedCompletion}(I,D,L_c)$
            \State $\mathcal{V}_{\mathrm{obl}}\gets \mathcal{V}_{\mathrm{obl}}\cup\{(I,D,\Gamma,\mathbf{K}_{ijk},\mathbf{E}_{ijk})\}$
        \EndFor
    \EndFor
\EndFor

\Statex
\State \textbf{Step 3: Generate sliding-window BEV views}
\State Generate BEV window origins $\mathcal{O}=\{(o_m^x,o_m^y)\}_{m=1}^{M}$ with tile size $G$ and stride $\Delta_g$
\ForAll{$\mathbf{o}_m=(o_m^x,o_m^y)\in\mathcal{O}$}
    \State $R_{\mathrm{bev}}\gets\lfloor G/s_g\rfloor$ \Comment{spatial resolution of BEV}
    \State Rasterize points into BEV RGB image $I^{\mathrm{bev}}$, height map $Z^{\mathrm{bev}}$, and correspondence map $\Gamma^{\mathrm{bev}}$
    \State Retain the topmost point if multiple points fall into the same BEV pixel
    \State $(I^{\mathrm{bev}},Z^{\mathrm{bev}})\gets \mathrm{DepthGuidedCompletion}(I^{\mathrm{bev}},Z^{\mathrm{bev}},L_c)$
    \State $\mathcal{V}_{\mathrm{bev}}\gets \mathcal{V}_{\mathrm{bev}}\cup\{(I^{\mathrm{bev}},Z^{\mathrm{bev}},\Gamma^{\mathrm{bev}},\mathbf{o}_m,s_g)\}$
\EndFor

\Statex
\State \textbf{Step 4: Return hybrid rendered views}
\State $\mathcal{V}\gets \mathcal{V}_{\mathrm{obl}}\cup\mathcal{V}_{\mathrm{bev}}$
\State \Return $\mathcal{V}$

\end{algorithmic}
\end{algorithm}

\subsubsection{Reliability-constrained Open-vocabulary Pseudo-labeling}\label{sec:pseudo-label}
Pre-trained VLMs offer an effective approach to open-vocabulary semantic segmentation. Given the hybrid rendered view set $\mathcal{V} = \mathcal{V}_{\mathrm{obl}}\cup\mathcal{V}_{\mathrm{bev}}$, COSTA queries the VLMs in a class-prompt manner. For each target class $c\in\mathcal{C}_t$, the VLMs return a probabilistic mask $S_{\nu,c}\in[0,1]^{H\times W}$ for the rendered view $\nu\in\mathcal{V}$. This prompt-conditioned inference allows COSTA to obtain semantic evidence for arbitrary user-specified classes, including categories beyond the source.

However, raw VLMs masks are not directly reliable as 3D pseudo labels. Due to significant domain gaps, occlusions, and poor rendering quality, these predictions are incomplete and inconsistent across views. Using them for direct propagation might be suboptimal. Therefore, we propose a reliability-aware pseudo-labeling strategy based on class-wise adaptive detection banks and multi-view evidence accumulation to suppress unreliable predictions naturally.

For each queried class $c$, a detection bank is initially constructed from mask responses larger than a low response threshold $\tau_0$, inspired by \cite{alami2026open}. The class-wise adaptive threshold is computed based on the statistical distribution as
\begin{equation}
\tau_c=\mu_c+a_c\sigma_c,
\label{eq:adaptive_threshold}
\end{equation}
where $\mu_c^q$ and $\sigma_c^q$ denote the mean and standard deviation of valid mask responses collected from the rendered view set $\mathcal{V}$. By default, we set $a_c^q=0$, reducing the threshold to the class-wise mean. This class-adaptive strategy avoids using a fixed global threshold, which is often suboptimal, since the response magnitudes of VLMs vary significantly across different classes.

After thresholding, COSTA back-projects the 2D masks to 3D points utilizing the established point-image correspondences. For a visible point $p_i$, if the mask response for class $c$ exceeds the adaptive threshold $\tau_c$, it converts into normalized semantic evidence for a specific view:
\begin{equation}
e_{i,\nu,c}
=
\begin{cases}
\dfrac{S_{\nu,c}(\pi_\nu(i))-\tau_c}{1-\tau_c},
& S_{\nu,c}(\pi_\nu(i))\ge\tau_c,\\
0, & \mathrm{otherwise}.
\end{cases}
\label{eq:evidence}
\end{equation}
Here, $\pi_\nu(i)$ denotes the projected pixel coordinate of the point $p_i$ in view $\nu$. The point-wise evidence for class $c$ is then aggregated by averaging all valid observations:
\begin{equation}
s_{i,c}
=
\frac{1}{|\mathcal{V}_i|}
\sum_{\nu\in\mathcal{V}_i}
e_{i,\nu,c}.
\label{eq:fused_evidence}
\end{equation}
where $\mathcal{V}_i\subseteq\mathcal{V}$ denotes the set of views where $p_i$ is visible. The preliminary pseudo label is determined by the class exhibiting the maximum aggregated evidence. To prevent erroneous forced assignments, points lacking valid evidence or possessing insufficient confidence are explicitly assigned an unknown label.

To select trustworthy semantic seeds, COSTA further computes a reliability score $r_i$ for each pseudo-labeled point. The score is formulated by jointly considering five cues: fused confidence, top-1/top-2 margin, evidence purity, cross-source agreement, and observation support. Let $\alpha_i$ and $\beta_i$ denote the top-1 and top-2 highest class confidence, respectively. The top-1/top-2 margin, defined as $\Delta_i=\alpha_i-\beta_i$, quantifies the separability between the dominant class and its closest competitor.

The reliability score is computed as a weighted combination of these normalized cues:
\begin{equation}
r_i=
\left[
O_i
\left(
0.35\alpha_i+0.25M_i+0.20Q_i+0.20A_i
\right)
\right]^{\gamma}.
\label{eq:reliability}
\end{equation}
In this formulation, $O_i=\log(1+|\mathcal{V}_i|)/\log(1+\kappa)$ measures the observation support, with the scale factor $\kappa$ set to $6$ by default; $M_i=\Delta_i/\lambda_\Delta$ is the normalized margin score; $Q_i=\alpha_i/(\sum_c s_{i,c}$ measures the purity of the dominant evidence; $A_i=|\{\nu\in\mathcal{V}_i:\hat{y}_{i,\nu}=\tilde{y}_i\}|/(|\mathcal{V}_i|)$ measures multi-view agreement, $\hat{y}_{i,\nu}=\arg\max_c e_{i,\nu,c}$ is view-level prediction; and $\gamma$ is a sharpening exponent set to $1.35$ in this implementation. Unknown predictions always receive $r_i=0$. The output of this stage is thus a reliability-aware pseudo-labeled point cloud such that:
\begin{equation}
\tilde{\mathcal{P}}_t
=
\{(p_i,\tilde{y}_i,r_i)\}_{i=1}^{N},
\label{eq:pseudo_point_cloud}
\end{equation}
where $\tilde{y}_i\in\mathcal{C}_t\cup\{c_{\mathrm{unk}}\}$ and $r_i\in[0,1]$. This stage is not intended to produce the final segmentation result. Instead, it converts open-vocabulary VLMs masks into reliable-aware 3D semantic seeds to facilitate subsequent adaptation and cluster-level propagation.

\subsection{Learnable Parameter Adaptation with Differentiated Self-Supervised Learning}
\begin{figure}[htbp]
\centering
\includegraphics[width=0.96\textwidth]{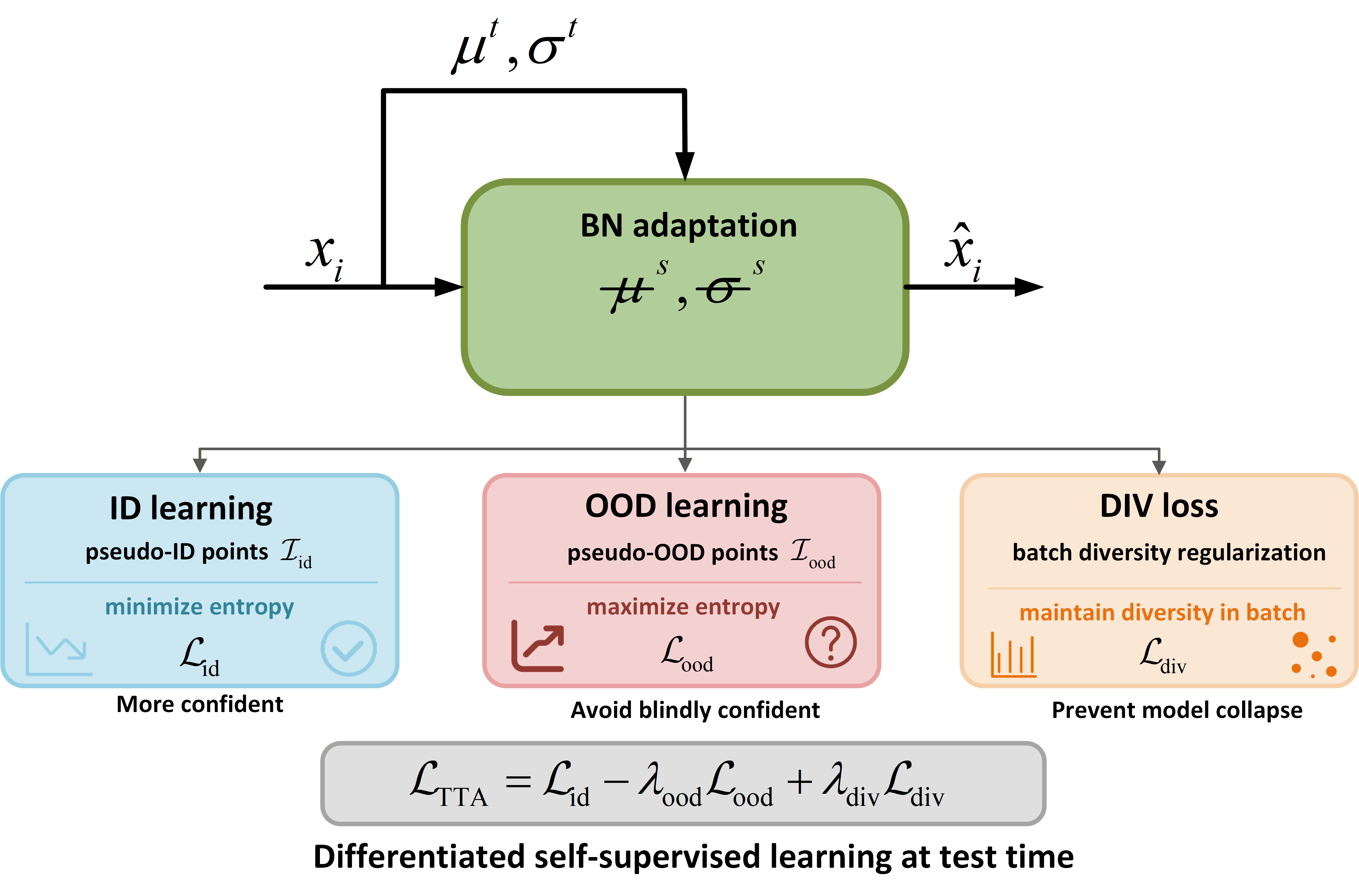}
\caption{Illustration of BN TTA with differentiated self-supervised learning. For each target batch, only the affine parameters $\{\gamma_l,\beta_l\}_{l\in\mathcal{L}_{\mathrm{BN}}}$ of the BN layers are optimized, while all remaining network parameters are frozen. Reliable pseudo-ID points are encouraged to produce confident source-space predictions through entropy minimization, whereas pseudo-OOD points are prevented from being over-confidently assigned to source classes through entropy maximization. A diversity regularization term further prevents class collapse and preserves batch-level discriminability.}\label{fig:TTA}
\end{figure}

Although VLMs provide open-vocabulary semantic evidence, they remain vulnerable to domain shifts. This issue is particularly pronounced in aerial point clouds due to severe modality and viewpoint discrepancies. Direct adaptation of VLMs entails optimizing a huge parameter space, making it an impractical solution. Therefore, we alternatively adopt a 3D segmentation model pre-trained on the source domain to exploit geometric features. Subsequently, we adapt the pre-trained 3D model to the target domain via a lightweight test-time adaptation strategy driven by differentiated self-supervised learning, where only BN layers are updated to capture domain-specific statistics (see Fig.~\ref{fig:TTA}).

The motivation is that batch normalization layers are highly sensitive to domain statistics~\citep{wang2021tent,wang2025test}. When the target distribution differs from the source distribution, the normalization statistics of the source domain can miscalibrate intermediate features. Updating only the affine parameters of BN provides a compact adaptation space, reducing the risk of overfitting noisy pseudo labels while effectively absorbing target-domain statistics. Formally, the learnable test-time parameters are as follows.
\begin{equation}
\Theta_{\mathrm{TTA}}
=
\{\gamma_l,\beta_l\}_{l\in\mathcal{L}_{\mathrm{BN}}},
\label{eq:bn_params}
\end{equation}
where $\gamma_l$ and $\beta_l$ denote the affine scale and shift of the $l$-th BN layer. All other parameters in the source model remain frozen during inference.

During inference, the adapted model generates source-space logits and decoder features for a given target batch. Since the source classifier is still defined over $\mathcal{C}_s$, COSTA does not attempt to directly classify new target classes in $\mathcal{C}_t$. Instead, we leverage the VLM-derived pseudo-labels to partition target points into in-distribution (ID) and out-of-distribution (OOD) sets. The assignment relies on the overlap between the target classes $\mathcal{C}_t$ and the source classes $\mathcal{C}_s$: a target category $c \in \mathcal{C}_t$ is regarded as ID if it also appears in $\mathcal{C}_s$, and as OOD otherwise. Formally,
\begin{equation}
\mathcal{C}_{\mathrm{id}} = \mathcal{C}_t \cap \mathcal{C}_s,\qquad
\mathcal{C}_{\mathrm{ood}} = (\mathcal{C}_t \setminus \mathcal{C}_s) \cup \{c_{\mathrm{unk}}\}.
\label{eq:id_ood_sets}
\end{equation}
Based on the pseudo label $\tilde{y}_i$, the target points are partitioned as pseudo-ID and pseudo-OOD samples.
\begin{equation}
\mathcal{I}_{\mathrm{id}}
=
\{i\mid \tilde{y}_i\in\mathcal{C}_{\mathrm{id}}\},
\quad
\mathcal{I}_{\mathrm{ood}}
=
\{i\mid \tilde{y}_i\in\mathcal{C}_{\mathrm{ood}}\}.
\label{eq:id_ood_split}
\end{equation} 
This design makes the open-set assignment fully controllable: by specifying $\mathcal{C}_t$, the user defines the ID/OOD split through the intersection with $\mathcal{C}_s$.

Inspired by \cite{gao2024unified}, COSTA follows the principle of unified entropy optimization for open-set test-time adaptation. Let $\mathbf{z}_i$ be the source-space logit of point $p_i$, and let
\begin{equation}
p_i(c)=
\frac{\exp(z_i^c)}
{\sum_{c\in\mathcal{C}_s}\exp(z_i^{c})},
\quad c\in\mathcal{C}_s,
\label{eq:source_prob}
\end{equation}
denote the source classifier probability. For pseudo-ID points, COSTA minimizes prediction entropy to encourage confident decisions for source-explainable semantics:
\begin{equation}
\mathcal{L}_{\mathrm{id}}
=
\frac{1}{|\mathcal{I}_{\mathrm{id}}|}
\sum_{i\in\mathcal{I}_{\mathrm{id}}}
H(p_i),
\quad
H(p_i)=
-\sum_{c\in\mathcal{C}_s}
p_i(c)\log p_i(c).
\label{eq:id_entropy}
\end{equation}
This term sharpens the source classifier response on target points that align with the source label space, thereby calibrating the target-domain features.

Conversely, for pseudo-OOD points, the objective is inverted. Since these points correspond to novel or target-specific semantics, forcing them into the source classes would destroy the model confidence. COSTA therefore maximizes their entropy over the closed-set source classifier:
\begin{equation}
\mathcal{L}_{\mathrm{ood}}
=
\frac{1}{|\mathcal{I}_{\mathrm{ood}}|}
\sum_{i\in\mathcal{I}_{\mathrm{ood}}}
H(p_i).
\label{eq:ood_entropy}
\end{equation}
By subtracting this OOD entropy term in the final minimization objective, the model is penalized for producing over-confident source-class predictions on target-specific classes. Crucially, unlike conventional methods that infer OOD samples solely from intrinsic classifier uncertainty, our OOD partition is explicitly guided by VLMs-derived open-vocabulary pseudo-labels.

Furthermore, COSTA incorporates a diversity regularization term to circumvent degenerate solutions. Entropy minimization may collapse most target points into a few dominant source classes. To preserve batch-level discriminability, the marginal prediction distribution is encouraged to remain diverse. Given the batch-averaged source prediction,
\begin{equation}
\bar{p}(c)=\frac{1}{N}\sum_{i=1}^{N}p_i(c),
\label{eq:marginal_prob}
\end{equation}
COSTA minimizes the negative entropy of this marginal distribution:
\begin{equation}
\mathcal{L}_{\mathrm{div}}
=
-\sum_{c\in\mathcal{C}_s}
\bar{p}(c)\log\left(\bar{p}(c)\right),
\label{eq:div_loss}
\end{equation}
Minimizing $\mathcal{L}_{\mathrm{div}}$ is equivalent to maximizing the marginal entropy, which prevents class collapse and ensures separable adapted features.

The final test-time adaptation objective is formulated as
\begin{equation}
\mathcal{L}_{\mathrm{TTA}}
=
\mathcal{L}_{\mathrm{id}}
-
\lambda_{\mathrm{ood}}\mathcal{L}_{\mathrm{ood}}
+
\lambda_{\mathrm{div}}\mathcal{L}_{\mathrm{div}}.
\label{eq:tta_loss}
\end{equation}
Only $\Theta_{\mathrm{TTA}}$ in Eq.~\eqref{eq:bn_params} is optimized by back-propagating Eq.~\eqref{eq:tta_loss}. Note that this stage is not intended to generate the final open-set labels. The classifier head still operates in the source label space $\mathcal{C}_s$. It plays a key role in the reshaping of the target-domain feature distribution so that semantically analogous points cluster compactly in the decoder feature space. The final open-set predictions are subsequently produced by the cluster-centric semantic propagation detailed in the next section.

\subsection{Cluster-Centric Open-Set Semantic Propagation}
This stage leverages the cluster-centric semantic propagation mechanism to integrate the domain generalization strengths of the TTA model with the open-vocabulary segmentation capability of VLMs. As shown in Fig.~\ref{fig:cluster_purity_intro}, we empirically found that target-domain points that share similar semantics tend to form compact clusters within the adapted decoder feature space following proven test-time adaptation. Intuitively, this property provides a natural mechanism to mitigate the noise inherent in VLMs-derived pseudo-labels. Rather than independently assigning labels to individual points, COSTA conceptualizes each cluster as a local semantic unit and performs semantic propagation at the cluster level.

Given the adapted decoder feature $\mathbf{h}_i$ of point $p_i$, COSTA first normalizes the features followed by K-Means clustering:
\begin{equation}
\bar{\mathbf{h}}_i=
\frac{\mathbf{h}_i}{\|\mathbf{h}_i\|_2},
\quad
\{\boldsymbol{\mu}_j\}_{j=1}^{J},\;a_i
=
\mathrm{KMeans}(\{\bar{\mathbf{h}}_i\}_{i=1}^{N}),
\label{eq:kmeans}
\end{equation}
where $\boldsymbol{\mu}_j$ represents the center of the $j$-th cluster and $a_i$ indicates the cluster assignment of the point $p_i$. The number of clusters is capped by the batch size, e.g., $J=\min(100,N)$. Each cluster is expected to contain points with similar semantics.

Given the inherent noise in pseudo-labels generated by VLMs, we avoid using all points within a cluster for equal propagation. Instead, it constructs a reliability-filtered semantic bank inside each cluster. A point is admitted to the cluster bank $\mathcal{B}_j$ only if its pseudo-label is valid and its reliability score exceeds a threshold $\rho_c$. These reliability score is derived from the pseudo-labeling stage (Sec.~\ref{sec:pseudo-label}). Formally, for $j$-th cluster $C_j$, this filtering could be formulated as
\begin{equation}
\mathcal{B}_j
=
\{p_i\in C_j\mid
\tilde{y}_i\in\mathcal{C}_t,\;
r_i\ge \rho_c
\},
\label{eq:cluster_bank}
\end{equation}
This filtering is important because various classes may exhibit varying response quality of VLMs, projection visibility, and spatial compactness.

During inference, we propagate the prediction of the cluster-level weighted voting to all points in each cluster. Unlike nearest-center propagation, we aggregate all reliable seeds in $\mathcal{B}_j$ within the group using the reliability-weighted strategy, which ensures that high-confidence seeds contribute more significantly than low-confidence seeds. The semantic prediction of each cluster is formulated as
\begin{equation}
\ell_j
=
\arg\max_{c\in\mathcal{C}_t}
\sum_{i\in\mathcal{B}_j}
r_i\,\mathbb{I}[\tilde{y}_i=c].
\label{eq:weighted_cluster_vote}
\end{equation}
where $\mathbb{I}$ denotes the indicator function. If $\mathcal{B}_j$ is empty, the cluster is temporarily assigned to the unknown label $c_{\mathrm{unk}}$. Otherwise, the class that accumulates the evidence with the highest reliability is designated as the cluster label. The final point-wise prediction is obtained by propagating the cluster label to all constituent  points in the cluster.

\section{Experiments}\label{sec4}
In this section, we first describe our experimental setup for OSTTA (Sec.~\ref{sec:protocol}). Then, we demonstrate the generality of our approach through cross-dataset evaluation (Sec.~\ref{sec:results}). Finally, we discuss and ablate our methods (Sec.~\ref{sec:analysis}).

\subsection{Evaluation Protocol of OSTTA}\label{sec:protocol}
\subsubsection{Benchmark}
We set up an evaluation protocol to simulate the conditions that occur when a pretrained aerial point cloud segmentation model is deployed to unseen scenes. In such applications, the source data is collected in a limited set of regions and annotated with source domain-specific classes. The deployed model is expected to operate on previously unseen scenes (cross-geographic and cross-sensor) without accessing source data and target annotations, which is reasonable in practice when source data cannot be redistributed or stored due to privacy, licensing, or storage constraints, and rapid deployment to a new target domain is required. Importantly, unlike closed-set TTA protocols that retain only shared classes or merge similar fine-grained classes for evaluation, we retain the native target classes to simulate the segmentation demanded by users. To study OSTTA in such a setting, we base our experiment setup on $4$ public aerial point cloud datasets, including DALES~\citep{varney2020dales}, SUM~\citep{sum2021}, Hessigheim 3D~\citep{kolle2021hessigheim}, and CUS3D~\citep{gao2024cus3d}. The characteristics of these datasets are described below.

DALES. Dayton Annotated Laser Earth Scan (DALES) is a large-scale benchmark dataset for 3D scene understanding. It consists of 40 airborne LiDAR tiles, each covering approximately 500m × 500m, collected in urban and suburban areas of the city of Surrey in British Columbia, Canada. The dataset provides dense per-point semantic labels across 9 classes.

SUM. Semantic Urban Mesh (SUM) is a photogrammetric benchmark dataset designed for semantic segmentation of urban environments, covering approximately 4 km² in Helsinki, Finland. The dataset is divided into 64 equally sized areas with a size of 250m x 250m and encompasses six annotated classes.

Hessigheim 3D. Hessigheim 3D (H3D) is a high-density airborne LiDAR point cloud dataset of approximately 800 points/m2. The dataset is captured by a Riegl VUX-1LR Scanner integrated with RGB images with a ground sampling distance (GSD) of 2 to 3 centimeters in the village of Hessigheim, Germany. A total of 11 fine-grained classes have been manually annotated. Here, we use test data collected in March 2018 for a standardized comparative evaluation.

CUS3D. CUS3D is a comprehensive urban-scale semantic segmentation benchmark dataset built upon large-scale 3D reconstruction using drone aerial imagery. It provides both color point clouds and textured mesh 3D data. The dataset covers approximately 2.85 km² of urban, suburban, and rural scenes, containing over 152 million 3D points. For mixed urban and rural scenes, each 3D point is annotated with one of 10 semantic classes.

We therefore use DALES as the labeled source domain and construct 3 OSTTA benchmarks, namely DALES-SUM, DALES-H3D, and DALES-CUS3D. The  detailed semantic relationships of 3 constructed OSTTA benchmarks are provided in Table.~\ref{tab:benchmark_mapping}. These benchmarks cover two representative semantic shifts: unseen classes and cross-granularity classes. Specifically, DALES-SUM mainly evaluates open-set adaptation to newly introduced target classes, where \cls{water} and \cls{boat} have no reliable counterparts in the source. DALES-H3D focuses on semantic granularity discrepancies, where most target classes are finer-grained or structurally redefined variants of source-domain semantics, such as vegetation-related and building-related subclasses. DALES-CUS3D provides a more comprehensive setting that simultaneously involves both semantic shift patterns.

\begin{table}[t]
\caption{Semantic relationship of the constructed OSTTA benchmarks between the target domain and the source domain. }
\label{tab:benchmark_mapping}
\centering
\small
\begin{tabularx}{\linewidth}{@{}lXXX@{}}
\toprule
Benchmark 
& Known classes 
& Cross-granularity classes 
& Unseen classes \\
\midrule
\bench{DALES-SUM} 
& \cls{terrain}; \cls{building}; \cls{car}
& \cls{high vegetation}
& \cls{water}; \cls{boat} \\

\bench{DALES-H3D} 
& \cls{vehicle}
& \cls{low vegetation}; \cls{shrub}; \cls{tree}; 
  \cls{impervious surface}; \cls{soil}; 
  \cls{roof}; \cls{facade}; \cls{chimney}; 
  \cls{vertical surface}
& \cls{/} \\

\bench{DALES-CUS3D} 
& \cls{building}; \cls{car};
& \cls{ground}; \cls{road}; 
  \cls{grass}; \cls{high vegetation}
& \cls{playground}; \cls{water}; \cls{farmland}; \cls{building site} \\
\botrule
\end{tabularx}
\footnotetext{Known classes denote target classes that directly overlap with the source classes. 
Cross-granularity classes denote target classes that are finer or coarser semantic variants of the source classes. 
Unseen classes denote target classes without reliable source-domain counterparts.}
\end{table}

\subsubsection{Metrics}
To comprehensively evaluate the performance of our approach, we evaluate semantic segmentation performance using per-class intersection over inion (IoU) and three overall metrics: overall accuracy (OA), mean intersection over union (mIoU), and mean class accuracy (mAcc). Let $\mathcal{C}$ denote the set of evaluated classes. For each class $c\in\mathcal{C}$, we compute the true positives $TP_c$, false positives $FP_c$, and false negatives $FN_c$ from the prediction results. The IoU of class $c$ is defined as
\begin{equation}
\mathrm{IoU}_c =
\frac{TP_c}{TP_c+FP_c+FN_c}.
\label{eq}
\end{equation}
The mean IoU is then obtained by averaging over all evaluated classes:
\begin{equation}
\mathrm{mIoU} =
\frac{1}{|\mathcal{C}|}
\sum_{c\in\mathcal{C}}
\mathrm{IoU}_c.
\label{eq:miou}
\end{equation}

In addition, we report overall accuracy, which measures the proportion of correctly classified points among all valid points:
\begin{equation}
\mathrm{OA} =
\frac{
\sum_{c\in\mathcal{C}} TP_c
}{
\sum{c\in\mathcal{C}} (TP_c+FN_c)
}.
\label{eq}
\end{equation}
To assess the average accuracy across classes, especially under class imbalance, we further report mean class accuracy:
\begin{equation}
\mathrm{mAcc} =
\frac{1}{|\mathcal{C}|}
\sum_{c\in\mathcal{C}}
\frac{TP_c}{TP_c+FN_c}.
\label{eq}
\end{equation}

\subsubsection{Implementation details}
We note that all experiments are run on NVIDIA RTX PRO 6000 GPUs and Linux OS with Ubuntu 22.04, AMD EPYC 9V74 CPU. We use KPConv~\citep{thomas2019kpconv} as the lightweight 3D segmentation model and a frozen Sa2VA~\citep{yuan2025sa2va} as the 2D VLM. The 3D model is pre-trained solely on the source domain, DALES. During training, we standardize the grid size to 0.25 m to subsample the original point cloud and improve computational efficiency, and set the overlapping spherical sub-cloud sampling to $15$ m for batch creation. Once pretraining on the source domain was complete, we deployed the pretrained model for semantic segmentation on the target test dataset during the inference stage, without accessing the source data and target annotations. Due to scale differences between datasets, we set overlap spherical sampling radius of 20 m, 15 m, and 15 m for SUM, H3D, and CUS3D, respectively, during the inference stage. All models are optimized by Adam optimizer with a learning rate of $10^{-4}$ and are implemented using the PyTorch framework.

For oblique virtual-camera projection, we set $K_l=6$ and place $3$ cameras around each anchor, yielding $6^{2}\times3=108$ candidate oblique virtual-camera views per point cloud. For BEV projection, we set $s_g=0.05$, $G=50\mathrm{m}, \Delta_g=25$ for SUM and $G=25\mathrm{m}, \Delta_g=12.5$ for H3D/CUS3D. For pseudo-labeling, Sa2VA was queried using the prompt template “Please segment all \(\{category\}\).” Mask responses below \(\tau_0=0.1\) were excluded. During cluster-centric propagation, every spherical sub-cloud was partitioned into $200$ clusters using \(k\)-means++ with a random seed of 0. In addition, $\rho_c$ in Eq.~(\ref{eq:cluster_bank}) is set as $0.4$.

\subsection{Results}\label{sec:results}

\subsubsection{Quantitative results}
In this section, we evaluate our approach under the OSTTA setup using the evaluation protocol described above. To comprehensively evaluate performance, we compare it with three types of baselines. As summarized in Table~\ref{tab:setting_comparison}, the compared methods follow different supervision and adaptation protocols. Fully-supervised methods are trained with target-domain annotations, open-vocabulary distillation methods require target-domain offline training, and multi-dataset pretraining methods rely on massive annotated datasets for foundation model construction. In contrast, COSTA follows a stricter open-set test-time adaptation setting: the source-pretrained model is directly deployed on unlabeled target data, without source-data replay, target annotations, or offline retraining. Therefore, the results should be interpreted not only by absolute performance, but also by the amount of supervision and training cost required by each setting. We provide the quantitative results on the three constructed OSTTA benchmarks in Table~\ref{tab:dales_sum_results}--\ref{tab:dales_cus3d_results_sideways}.

\textbf{DALES-SUM.} The DALES-SUM benchmark mainly evaluates adaptation performance to unseen target classes. As shown in Table~\ref{tab:dales_sum_results}, COSTA achieves $70.09\%$ mIoU and $88.52\%$ OA under the strict open-set TTA setting, which outperforms most baselines. Among fully-supervised methods, Point Transformer V3 achieves the best performance with $74.00\%$ mIoU and $94.70\%$ OA, benefiting from its transformer-based architecture and target-domain supervised training. In contrast, COSTA obtains a comparable mIoU without any target annotation and offline re-training. In particular, our method surpasses representative supervised baselines such as SparseConv and KPConv by $27.49\%$ and $1.29\%$ mIoU, respectively. Compared with open-vocabulary distillation methods, COSTA yields a performance on par with OpenUrban3D, with merely a $5.31\%$ mIoU gap, while requiring no target-domain distillation and offline re-training. COSTA outperforms OpenScene by $28.99\%$ mIoU and $13.92\%$ OA. We attribute this to the limitation that direct open-vocabulary feature transfer is insufficient under domain and semantic shifts. For multi-dataset pretraining methods, it should be noted that this setting follows a closed-set evaluation protocol, where models are pretrained on multiple datasets including the train-set of target dataset before testing. Even under this favorable setting, COSTA is only $1.61\%$ mIoU lower than previous SoTA CitySeg-FM and exceeds Sonata by $8.89\%$ mIoU and $18.22\%$ OA. For unseen target classes, COSTA improves \cls{boat} from $61.32\%$ to $77.99\%$ and \cls{water} from $85.83\%$ to $88.16\%$ over Sa2VA + hybrid, and further outperforms OpenUrban3D on \cls{boat} by $9.89\%$ IoU. The results demonstrate that our approach can effectively alleviate domain shift in known classes and successfully discover unseen target-specific classes beyond the source.

\textbf{DALES-H3D.} The DALES-H3D benchmark introduces the challenge of cross-granularity semantic shifts, where most target classes are not strictly unseen but correspond to finer-grained or structurally redefined variants of source-domain semantics, such as vegetation-related subclasses and building-related subclasses. As shown in Table~\ref{tab:dales_h3d_results_sideways}, COSTA achieves $34.13\%$ mIoU and $70.68\%$ OA. Compared with fully-supervised methods, our approach operates under the OSTTA setting, and thus still exhibits a clear gap from strong fully-supervised models on this fine-grained benchmark. Nevertheless, COSTA exceeds PointNet++ by $4.23\%$ mIoU and achieves competitive OA compared to early supervised baselines such as PointNet and PointNet++. Compared with multi-dataset pretraining methods, COSTA outperforms Sonata by $0.33\%$ mIoU and $7.58\%$ OA, and surpasses PTv3 + PPT by $3.48\%$ OA, although it remains $17.77\%$ mIoU lower than SoTA CitySeg-FM. This indicates that large-scale multi-dataset pretraining is still advantageous for dense fine-grained target classes, while COSTA provides a training-free alternative that can be deployed directly at test time. At the class level, COSTA performs reasonably well on \cls{low vegetation}, \cls{impervious surface}, \cls{vehicle}, \cls{roof}, \cls{tree}, and especially \cls{soil/gravel}, where it achieves $51.58\%$ IoU and even surpasses the best fully-supervised result by $8.13\%$. However, its performance remains limited on highly fine-grained, sparse, and small scale classes such as \cls{urban furniture} and \cls{chimney}. These results suggest that cross-granularity OSTTA is substantially more challenging than unseen-class adaptation because it requires not only recognizing target-specific semantics, but also separating subtle semantic subdivisions that are absent in the source. Furthermore, poor image rendering quality is also a significant constraint that hinders tiny object recognition.

\textbf{DALES-CUS3D.} The DALES-CUS3D benchmark provides a more comprehensive OSTTA setting that jointly involves unseen classes and cross-granularity semantic shifts. As shown in Table~\ref{tab:dales_cus3d_results_sideways}, COSTA achieves $49.30\%$ mIoU and $77.98\%$ OA under the open-set TTA setting. Among all fully-supervised methods, KPConv obtains the best overall performance with $59.72\%$ mIoU and $89.42\%$ OA with dense target-domain supervision. While there is $10.42\%$ mIoU gap against the best fully-supervised method (KPConv) due to the absence of target annotations and offline training, our approach outperforms representative supervised baselines like SPGraph by $18.86\%$ mIoU, and achieves competitive performance compared with RandLA-Net and PointNet++. Compared with the open-vocabulary baseline Sa2VA, COSTA improves mIoU from $39.28\%$ to $49.30\%$ and OA from $61.53\%$ to $77.98\%$. This indicates that directly transferring VLM-derived pseudo labels is insufficient for reliable 3D segmentation due to significant domain shift, while our approach can substantially improve 3D semantic consistency without additional model retraining. At the class level, COSTA brings clear improvements over Sa2VA on both unseen and cross-granularity classes. For unseen classes, it improves \cls{playground} from $6.24\%$ to $32.53\%$, \cls{farmland} from $53.92\%$ to $63.06\%$, \cls{building site} from $3.68\%$ to $11.47\%$, and \cls{water} from $78.06\%$ to $79.83\%$. For cross-granularity classes, it improves \cls{high vegetation} from $53.70\%$ to $74.99\%$ and \cls{grass} from $33.71\%$ to $38.87\%$, while maintaining comparable performance on \cls{road}. Notably, COSTA also achieves strong results on \cls{car} and \cls{water}, reaching $53.58\%$ and $79.83\%$ IoU, respectively. These results demonstrate that COSTA can handle mixed semantic shifts more effectively than direct open-vocabulary methods, especially for semantically well-defined and large-scale objects.

\begin{table}[t]
\centering
\caption{Comparison of different segmentation and adaptation settings.}
\label{tab:setting_comparison}
\renewcommand{\arraystretch}{1.15}
\begin{tabular}{lccc}
\toprule
\textbf{Setting} 
& \textbf{Pretrained model} 
& \textbf{Target data} 
& \textbf{Offline training} \\
\midrule
Fully-supervised method 
& $\times$
& $x^{t}, y^{t}$ 
& \checkmark \\

Open-vocabulary distillation 
& \checkmark
& $x^{t}$ 
& \checkmark \\

Multi-dataset pretraining
& \checkmark
& $x^{t}, y^{t}$ 
& \checkmark \\

Open-set test-time adaptation 
& \checkmark
& $x^{t}$ 
& $\times$ \\
\bottomrule
\end{tabular}
\end{table}

\begin{table*}[t]
\centering
\caption{Quantitative comparison on the DALES-SUM benchmark. 
Target classes are divided into known, cross-granularity, and unseen classes according to their semantic relationship with the source.}
\label{tab:dales_sum_results}
\resizebox{\textwidth}{!}{
\begin{tabular}{lccKCKUKU}
\toprule
\textbf{Method} 
& \textbf{mIoU} 
& \textbf{OA} 
& \textbf{Terrain} 
& \textbf{High Veg.} 
& \textbf{Building} 
& \textbf{Water} 
& \textbf{Vehicle} 
& \textbf{Boat} \\
\midrule

\rowcolor{gray!20}
\multicolumn{9}{l}{\textbf{Fully-supervised}} \\
PointNet~\citep{qi2017pointnet}              & 23.70 & 80.70 & 67.90 & 89.50 & 80.00 & 0.00  & 0.00  & 3.90  \\
PointNet++~\citep{qi2017pointnet}            & 32.90 & 84.30 & 72.40 & 94.20 & 84.70 & 2.70  & 2.00  & 25.70 \\
SPGraph~\citep{landrieu2018large}               & 37.20 & 76.90 & 69.90 & 94.50 & 88.80 & 32.80 & 12.50 & 15.70 \\
SparseConv~\citep{graham20183d}            & 42.60 & 85.20 & 74.10 & \textbf{97.90} & \textbf{94.20} & 63.30 & 7.50  & 24.20 \\
MinkUnet~\citep{choy20194d}              & 70.00 & 93.30 & 84.40 & 92.00 & 92.90 & 76.80 & \textbf{61.60} & 12.60 \\
KPConv~\citep{thomas2019kpconv}                & 68.80 & 94.20 & 86.50 & 88.40 & 92.70 & 77.70 & 54.30 & 13.30 \\
RandLA-Net~\citep{hu2020randla}            & 38.60 & 74.90 & 38.90 & 59.60 & 81.50 & 27.70 & 22.00 & 2.10  \\
Point Transformer V2~\citep{wu2022point}  & 65.40 & 93.10 & 82.80 & 89.60 & 93.60 & 87.40 & 19.80 & 18.80 \\
Point Transformer V3~\citep{wu2024point}  & \textbf{74.00} & \textbf{94.70} & \textbf{87.30} & 92.30 & \textbf{94.20} & \textbf{88.50} & 49.40 & \textbf{32.40} \\
\midrule

\rowcolor{gray!20}
\multicolumn{9}{l}{\textbf{Open-vocabulary distillation}} \\
OpenScene~\citep{peng2023openscene}             & 41.10 & 74.60 & 35.60 & 54.10 & 73.70 & 53.90 & 0.10  & 28.80 \\
PLA~\citep{ding2023pla}                   & 4.30  & 26.70 & 24.60 & 0.00  & 0.20  & 0.00  & 0.90  & 0.00  \\
RegionPLC~\citep{yang2024regionplc}             & 8.10  & 26.40 & 16.20 & 7.50  & 19.30 & 3.10  & 0.10  & 2.60  \\
Sa2VA + hybrid~\citep{yuan2025sa2va}        & 62.89 & 81.07 & 63.17 & 63.10 & 78.19 & 85.83 & 25.75 & 61.32 \\
OpenUrban3D~\citep{wang2026openurban3d}           & \textbf{75.40} & \textbf{90.50} & \textbf{69.30} & \textbf{83.60} & \textbf{90.20} & \textbf{92.70} & \textbf{48.30} & \textbf{68.10} \\
\midrule

\rowcolor{gray!20}
\multicolumn{9}{l}{\textbf{Multi-dataset pretraining}} \\
PTv3 + PPT~\citep{wu2024towards}
& 53.8 & 63.7
& $-$ & $-$ & $-$
& $-$ & $-$ & $-$  \\

Sonata~\citep{wu2025sonata}
& 61.2 & 70.3
& $-$ & $-$ & $-$
& $-$ & $-$ & $-$  \\

CitySeg-FM~\citep{xu2025cityseg}
& \textbf{71.7} & \textbf{91.7}
& $-$ & $-$ & $-$
& $-$ & $-$ & $-$  \\
\midrule

\rowcolor{gray!20}
\multicolumn{9}{l}{\textbf{Open-set TTA}} \\
COSTA                 & \textbf{70.09} & \textbf{88.52} & \textbf{68.52} & \textbf{72.45} & \textbf{86.53} & \textbf{88.16} & \textbf{26.90} & \textbf{77.99} \\
\bottomrule
\end{tabular}
}
\vspace{0.5em}

\footnotesize
\colorbox{knowncolor}{\strut Known classes} \quad
\colorbox{crosscolor}{\strut Cross-granularity classes} \quad
\colorbox{unseencolor}{\strut Unseen classes}
\end{table*}

\begin{sidewaystable}[p]
\centering
\small
\setlength{\tabcolsep}{0.3pt}
\renewcommand{\arraystretch}{1.00}

\caption{Quantitative comparison on the DALES-to-H3D benchmark. Target classes are grouped into known, cross-granularity, and unseen classes according to their semantic relationship with the DALES source label space.}
\label{tab:dales_h3d_results_sideways}

\begingroup
\def\Hone#1{%
  \begin{tabular}[c]{@{}c@{}}%
  \rule[-1.0ex]{0pt}{4.0ex}\textbf{#1}%
  \end{tabular}%
}
\def\Htwo#1#2{%
  \begin{tabular}[c]{@{}c@{}}%
  \rule{0pt}{1.8ex}\textbf{#1}\\[0.15ex]%
  \rule{0pt}{1.8ex}\textbf{#2}%
  \end{tabular}%
}

\begin{tabular*}{1.00\textheight}{@{\extracolsep\fill}lccCCKCCCCCCCC}
\toprule
\Hone{Method}
& \Hone{mIoU}
& \Hone{OA}
& \Htwo{Low}{Veg.}
& \Htwo{Imp.}{Surface}
& \Hone{Vehicle}
& \Htwo{Urban}{Furn.}
& \Hone{Roof}
& \Hone{Façade}
& \Hone{Shrub}
& \Hone{Tree}
& \Htwo{Soil/}{Gravel}
& \Htwo{Vertical}{Surf.}
& \Hone{Chimney} \\
\midrule

\rowcolor{gray!20}
\multicolumn{14}{l}{\textbf{Fully-supervised}} \\
PointNet~\citep{qi2017pointnet} 
& 36.64 & 71.05
& 56.00 & 49.03 & 2.00
& 21.06 & 69.12 & 43.21 & 23.02 & 83.03
& 3.09 & 42.97 & 1.02 \\

PointNet++~\citep{qi2017pointnet} 
& 29.90 & 68.50
& 64.13 & 56.26 & 18.85
& 7.30 & 58.71 & 31.37 & 16.49 & 56.00
& 5.09 & 12.07 & 2.24 \\

RandLA-Net~\citep{hu2020randla}  
& 57.31 & 83.99
& 69.05 & 72.03 & 37.26
& 43.60 & 85.33 & 68.61 & 45.32 & 90.24
& 12.96 & 63.03 & 42.97 \\

PIIE-DSA-Net~\citep{gao2022piie}
& 52.48 & 84.20
& 78.09 & 74.86 & 35.51
& 22.53 & 90.55 & 53.00 & 31.12 & 89.42
& 14.37 & 49.22 & 38.59 \\

KPConv~\citep{thomas2019kpconv}  
& 63.20 & 87.69
& 79.54 & 80.11 & \textbf{69.72}
& 46.94 & \textbf{94.51} & \textbf{74.21} & \textbf{60.32} & \textbf{94.90}
& 27.09 & \textbf{67.87} & 0.00 \\

SCN~\citep{kolle2021hessigheim}
& 64.87 & 88.42
& 85.65 & 78.83 & 46.53
& 39.97 & 93.94 & 71.07 & 52.18 & 94.15
& 28.90 & 64.24 & \textbf{58.19} \\

Point Transformer V3~\citep{wu2024point}
& \textbf{70.08} & \textbf{92.81}
& \textbf{88.71} & \textbf{85.67} & 65.18
& \textbf{53.90} & 94.20 & 72.83 & 51.35 & 90.88
& \textbf{43.45} & 66.64 & 58.12 \\
\midrule

\rowcolor{gray!20}
\multicolumn{14}{l}{\textbf{Open-vocabulary distillation}} \\
Sa2VA + hybrid~\citep{yuan2025sa2va}        & \textbf{28.38} & \textbf{60.16} & \textbf{54.04} & \textbf{56.55} & \textbf{35.24} & \textbf{8.00} & \textbf{71.49} & \textbf{12.04} &\textbf{5.01} &\textbf{31.82} &\textbf{29.93} &\textbf{7.77} &\textbf{0.34} \\
\midrule

\rowcolor{gray!20}
\multicolumn{14}{l}{\textbf{Multi-dataset pretraining}} \\
PTv3 + PPT~\citep{wu2024towards}
& 35.70 & 67.20
& $-$ & $-$ & $-$
& $-$ & $-$ & $-$ & $-$ & $-$
& $-$ & $-$ & $-$ \\

Sonata~\citep{wu2025sonata}
& 33.80 & 63.10
& $-$ & $-$ & $-$
& $-$ & $-$ & $-$ & $-$ & $-$
& $-$ & $-$ & $-$ \\

CitySeg-FM~\citep{xu2025cityseg}
& \textbf{51.90} & \textbf{81.20}
& $-$ & $-$ & $-$
& $-$ & $-$ & $-$ & $-$ & $-$
& $-$ & $-$ & $-$ \\
\midrule

\rowcolor{gray!20}
\multicolumn{14}{l}{\textbf{Open-set TTA}} \\
COSTA
& \textbf{34.13} & \textbf{70.68}
& \textbf{61.63} & \textbf{61.53} & \textbf{47.54}
& \textbf{4.94} & \textbf{66.94} & \textbf{12.92} & \textbf{2.51} & \textbf{57.41}
& \textbf{51.58} & \textbf{8.44} & 0.00 \\
\botrule
\end{tabular*}
\endgroup

\footnotetext{
\colorbox{knowncolor}{\strut Known classes} \quad
\colorbox{crosscolor}{\strut Cross-granularity classes} \quad
\colorbox{unseencolor}{\strut Unseen classes}.}
\end{sidewaystable}

\begin{sidewaystable}[p]
\centering
\small
\setlength{\tabcolsep}{0.3pt}
\renewcommand{\arraystretch}{1.00}

\caption{Quantitative comparison on the DALES-to-CUS3D benchmark. Target classes are grouped into known, cross-granularity, and unseen classes according to their semantic relationship with the DALES source label space.}
\label{tab:dales_cus3d_results_sideways}

\begingroup
\def\Hone#1{%
  \begin{tabular}[c]{@{}c@{}}%
  \rule[-1.0ex]{0pt}{4.0ex}\textbf{#1}%
  \end{tabular}%
}
\def\Htwo#1#2{%
  \begin{tabular}[c]{@{}c@{}}%
  \rule{0pt}{1.8ex}\textbf{#1}\\[0.15ex]%
  \rule{0pt}{1.8ex}\textbf{#2}%
  \end{tabular}%
}

\begin{tabular*}{1.00\textheight}{@{\extracolsep\fill}lcccCKCCUKUUCU}
\toprule
\Hone{Method}
& \Hone{mIoU}
& \Hone{OA}
& \Hone{mAcc}
& \Hone{Ground}
& \Hone{Build.}
& \Hone{Road}
& \Htwo{High}{Veg.}
& \Hone{Water}
& \Hone{Car}
& \Hone{Play.}
& \Hone{Farm.}
& \Hone{Grass}
& \Htwo{Building}{Site} \\
\midrule

\rowcolor{gray!20}
\multicolumn{14}{l}{\textbf{Fully-supervised}} \\
PointNet~\citep{qi2017pointnet}  
& 42.54 & 74.57 & 94.91
& \textbf{78.36} & 65.81 & \textbf{78.34} & 73.32
& 25.22 & 20.35 & 8.76 & 8.02 & 39.11 & 16.66 \\

PointNet++~\citep{qi2017pointnet} 
& 55.16 & 84.52 & 96.90
& 25.78 & 80.36 & 57.77 & 86.18
& 57.52 & 36.13 & 11.38 & 40.88 & 40.17 & 13.28 \\

RandLA-Net~\citep{hu2020randla}   
& 52.98 & 77.12 & 95.42
& 15.29 & 76.18 & 49.52 & 78.42
& \textbf{62.30} & 49.89 & \textbf{35.91} & \textbf{42.11} & 39.74 & 29.86 \\

KPConv~\citep{thomas2019kpconv}   
& \textbf{59.72} & \textbf{89.42} & \textbf{97.88}
& 25.73 & \textbf{82.89} & 48.69 & \textbf{89.64}
& 48.43 & 44.69 & 25.72 & 41.86 & 44.84 & 15.50 \\

SPGraph~\citep{landrieu2018large}
& 30.44 & 67.23 & 40.73
& 12.05 & 63.40 & 30.56 & 63.40
& 31.71 & 4.40 & 7.65 & 36.66 & 31.36 & 28.84 \\

SQN~\citep{hu2022sqn}
& 56.82 & 78.19 & 95.64
& 30.65 & 79.08 & 60.22 & 79.08
& 59.21 & \textbf{55.46} & 34.77 & 39.88 & \textbf{45.54} & \textbf{30.73} \\
\midrule

\rowcolor{gray!20}
\multicolumn{14}{l}{\textbf{Open-vocabulary distillation}} \\
Sa2VA+hybrid~\citep{yuan2025sa2va}
& \textbf{39.28} & \textbf{61.53} & \textbf{57.23}
& \textbf{8.02} & \textbf{58.74} & \textbf{57.58} & \textbf{53.70}
& \textbf{78.06} & \textbf{39.13} & \textbf{6.24} & \textbf{53.92} & \textbf{33.71} & \textbf{3.68} \\
\midrule

\rowcolor{gray!20}
\multicolumn{14}{l}{\textbf{Open-set test-time adaptation}} \\
COSTA
& \textbf{49.30} & \textbf{77.98} & \textbf{61.40}
& \textbf{7.81} & \textbf{72.78} & \textbf{58.07} & \textbf{74.99}
& \textbf{79.83} & \textbf{53.58} & \textbf{32.53} & \textbf{63.06} & \textbf{38.87} & \textbf{11.47} \\
\botrule
\end{tabular*}
\endgroup

\footnotetext{
\colorbox{knowncolor}{\strut Known classes} \quad
\colorbox{crosscolor}{\strut Cross-granularity classes} \quad
\colorbox{unseencolor}{\strut Unseen classes}.}
\end{sidewaystable}

\subsubsection{Qualitative results}
\begin{figure}[htbp]
\centering
\includegraphics[width=0.96\textwidth]{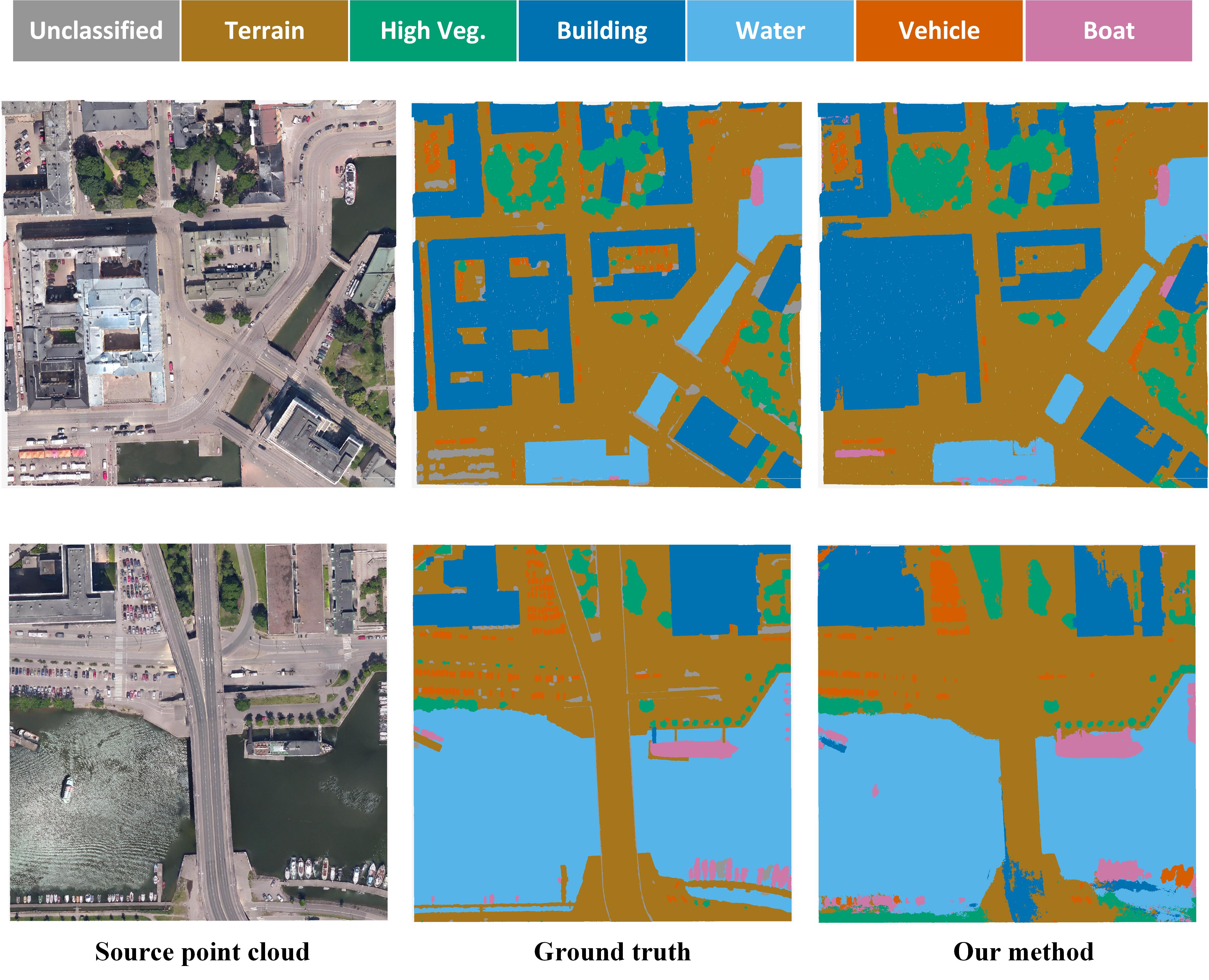}
\caption{Classification result of SUM dataset with a model adapted from DALES dataset.}\label{fig:SUM}
\end{figure}

\begin{figure}[htbp]
\centering
\includegraphics[width=0.96\textwidth]{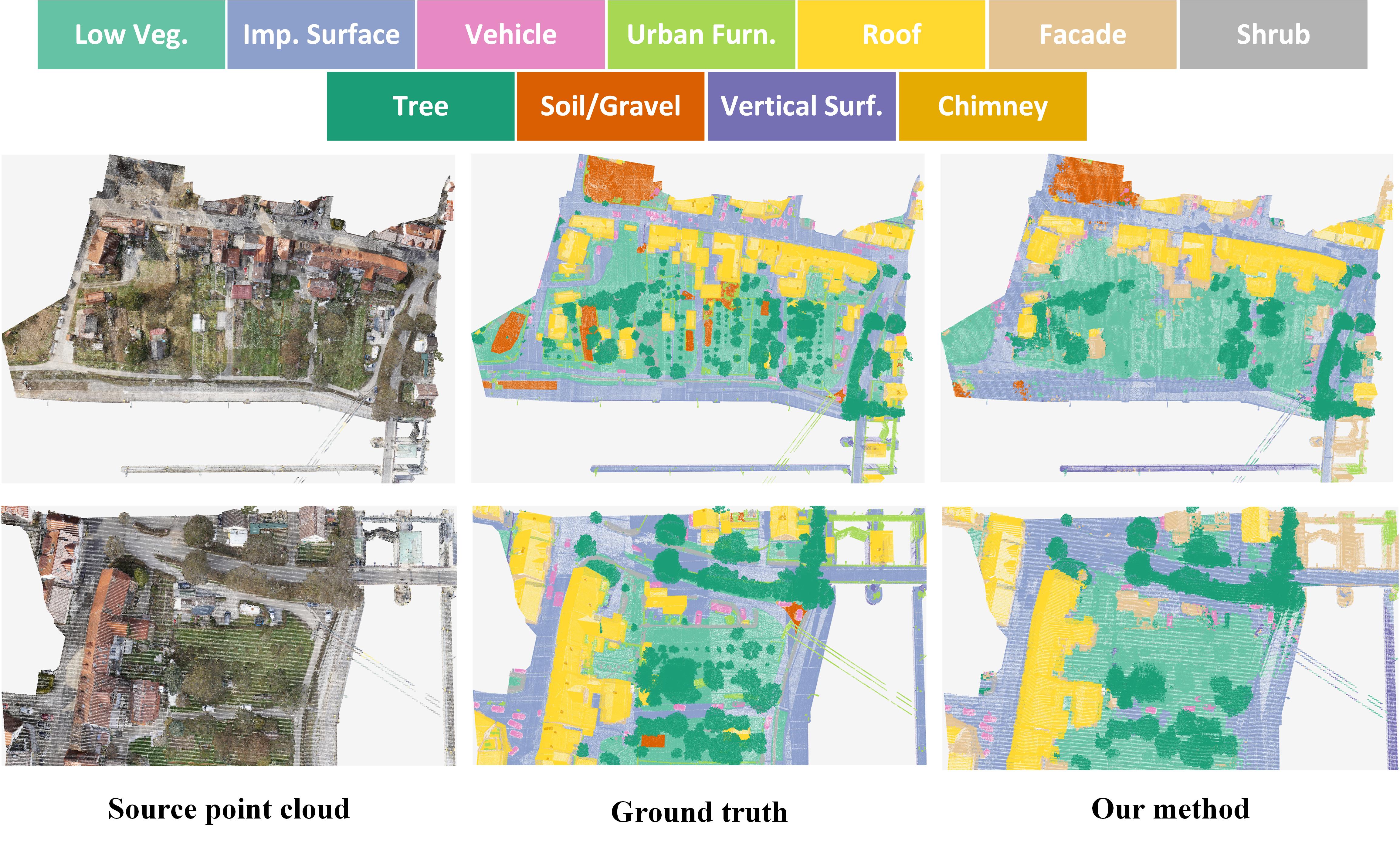}
\caption{Classification result of H3D dataset with a model adapted from DALES dataset.}\label{fig:H3D}
\end{figure}

\begin{figure}[htbp]
\centering
\includegraphics[width=0.96\textwidth]{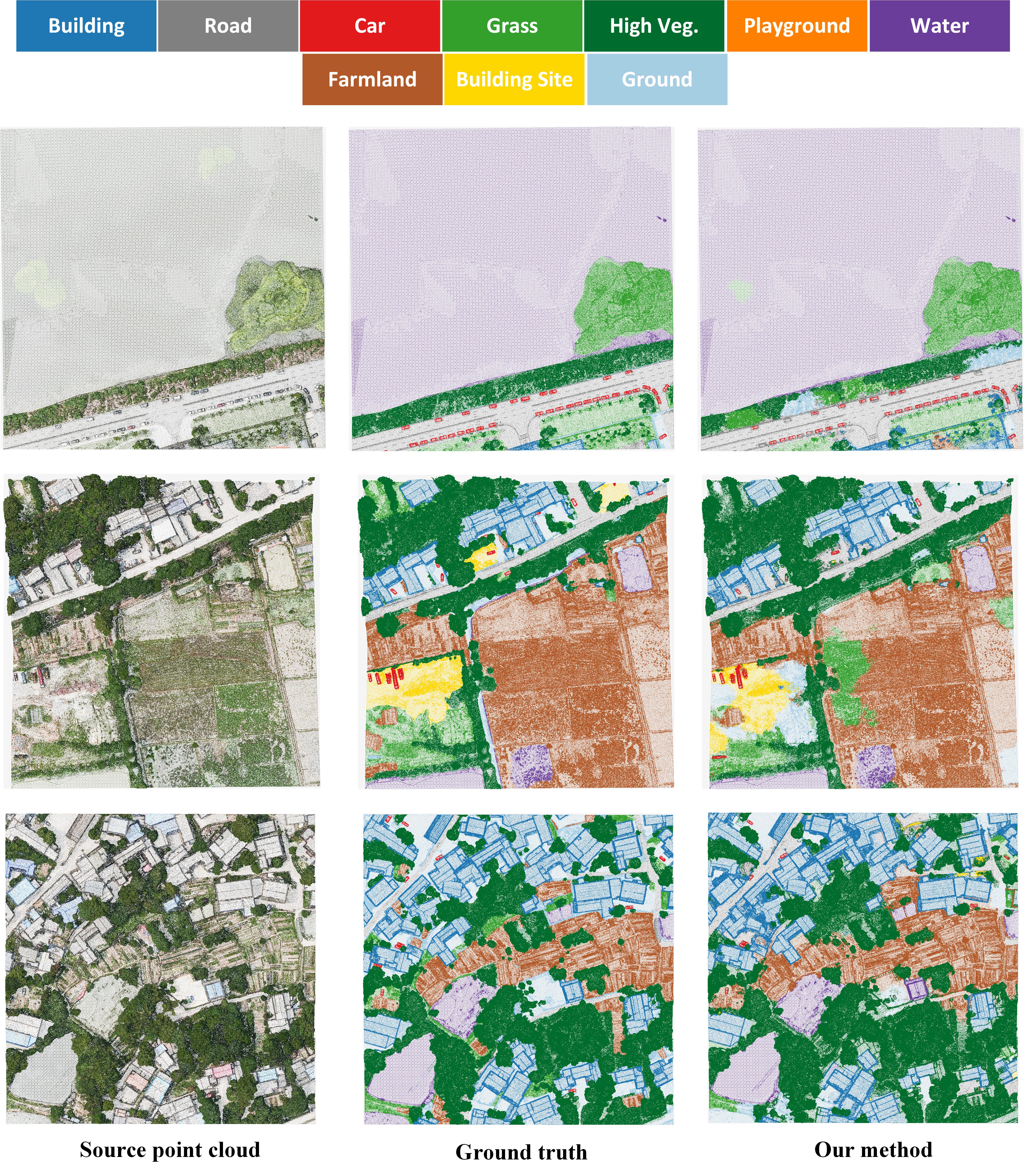}
\caption{Classification result of CUS3D dataset with a model adapted from DALES dataset.}\label{fig:CUS3D}
\end{figure}

In Fig.~\ref{fig:SUM}--\ref{fig:CUS3D}, we show several visual examples for COSTA in the three benchmarks. As can be seen, COSTA performs well in challenging cases with known classes and unseen classes in various scenes. For example, in the DALES-SUM benchmark, our method segments the known classes under covariate shifts, such as \cls{terrain}, \cls{high Veg.} and \cls{building}. Moreover, COSTA segments the unseen classes under semantic shifts, such as \cls{water} and \cls{boat}. This is especially true for \cls{water} that is relatively large in scale and has clearly defined categories. Similar observations could also be found in the DALES-H3D (see Fig.~\ref{fig:H3D}) and DALES-CUS3D (see Fig.~\ref{fig:CUS3D}) benchmark. Our method successfully segments \cls{impervious surfaces}, \cls{roofs}, \cls{farmland}, etc.

However, we also observe failure cases. In Fig~\ref{fig:H3D}, our method struggles to segment too tiny objects and targets with unclear category definitions. For example, in the DALES-H3D benchmark, the performance of small-scale \cls{chimney} and \cls{urban furniture} mixing various types of urban infrastructure is not satisfactory.

\subsection{Analysis}\label{sec:analysis}

\subsubsection{Analysis of hybrid multi-view projection}\label{sec:multi-view projection}
A prerequisite for good pseudo-labels in VLMs is high-quality multi-view projections. Here, we study the pseudo-label quality of our method with different multi-view projection strategies based on the same VLM~\citep{yuan2025sa2va}. Compared to the BEV-only projection, the oblique-only projection substantially improves OA, mAcc, and mIoU by $29.90\%$, $16.63\%$, and $15.86\%$ in the SUM dataset, while reducing the unknown ratio from $38.83\%$ to $8.26\%$ (Table~\ref{tab:pseudo_overall_modes}). Similar trends can also be observed in the H3D and CUS3D dataset. The oblique-only projection generally produces higher-quality pseudo labels. This is mainly because oblique virtual-camera views provide provide more comprehensive observations from diverse perspectives, which helps the 2D VLMs generate more complete prediction masks for reference. In contrast, BEV projection only provides a top-down observation. Although it offers a stable global layout, it tends to lose vertical and side-view information and is more sensitive to top-view occlusion, especially for structures such as building facades, vertical surfaces, and objects partially hidden by higher ones. As a result, BEV-only projection produces a larger proportion of unknown predictions.

The best performance is achieved when fusing both projection strategies, with a performance gain of approx. $+4$\%, $+1$\%, $+6$\% mIoU across SUM, H3D and CUS3D. Similar trends are observed for the mAcc and unknown ratio (Table~\ref{tab:pseudo_overall_modes}). The results demonstrate the effectiveness of the proposed hybrid multi-view evidence accumulation fusion strategy. By integrating complementary evidence from BEV and oblique virtual-camera projections, the fusion strategy can suppress unreliable predictions naturally, improve pseudo-label coverage, and generate more accurate pseudo labels.

The detailed per-class results (see Table~\ref{tab:sum_mode_per_class_iou}--Table~\ref{tab:cus3d_mode_per_class_iou}) provide further insight into the complementarity between BEV and oblique projections. BEV projection is beneficial for some compact or top-view distinguishable classes. For example, it achieves the best IoU for \cls{vehicle} on SUM and H3D, outperforming the oblique-only by approx. $10$\%. This is mainly because BEV provides a geometrically stable overhead footprint, which can effectively preserve the spatial layout of small objects that are visible from the top. However, BEV projection is less effective for classes that require oblique or side-view information. For example, its IoU on SUM \cls{building} is approximately $30\%$ lower than the oblique-only, and its IoU on H3D \cls{facade} and \cls{vertical surface} is also limited, indicating that top-down projection alone cannot sufficiently capture vertical structures. In contrast, oblique projection is more effective for large-scale and spatially extended categories, with a performance gain of approx. $+20$\% on \cls{water} of SUM.

Particularly, the fused strategy combines the advantages of both projection modes and achieves more balanced performance across categories. Compared to single projection strategies, the fused setting obtains the best IoU on \cls{ground}, \cls{high vegetation}, \cls{building}, and \cls{boat}, while maintaining competitive performance on \cls{water} and \cls{vehicle} on the SUM dataset. Similar results were observed on both the H3D and CUS3D datasets. The results confirm that BEV and oblique projections provide complementary semantic evidence: BEV captures stable global layout and top-view object footprints, whereas oblique projections offer diverse viewpoint-dependent details. Also, the multi-view evidence accumulation fusion strategy effectively combines the advantages of both projection. Overall, consistent improvements across various metrics demonstrate that the proposed hybrid projection and multi-view evidence accumulation fusion strategy can effectively enhance pseudo-label quality.

\begin{table}[t]
\centering
\caption{Overall pseudo-label evaluation results under different projection modes. 
The best OA, mAcc, and mIoU within each dataset are highlighted in bold, and the lowest unknown ratio is also highlighted.}
\label{tab:pseudo_overall_modes}
\renewcommand{\arraystretch}{1.12}
\setlength{\tabcolsep}{6pt}
\begin{tabular}{llcccc}
\toprule
\textbf{Dataset} & \textbf{Mode} & \textbf{OA (\%)} & \textbf{mAcc (\%)} & \textbf{mIoU (\%)} & \textbf{Unknown ratio (\%)} \\
\midrule
\multirow{3}{*}{SUM}
& BEV-only   & 48.82 & 51.20 & 43.06 & 38.83 \\
& oblique-only & 78.72 & 67.83 & 58.92 & 8.26  \\
& Fused      & \textbf{81.07} & \textbf{73.08} & \textbf{62.89} & \textbf{5.90} \\
\midrule
\multirow{3}{*}{H3D}
& BEV-only   & 48.94 & 33.18 & 24.17 & 20.61 \\
& oblique-only & \textbf{60.84} & 38.41 & 27.22 & 14.80 \\
& Fused      & 60.16 & \textbf{42.20} & \textbf{28.38} & \textbf{7.27} \\
\midrule
\multirow{3}{*}{CUS3D}
& BEV-only   & 46.87 & 50.97 & 33.12 & 26.72 \\
& oblique-only & 50.91 & 49.46 & 33.51 & 9.09  \\
& Fused      & \textbf{61.53} & \textbf{57.23} & \textbf{39.28} & \textbf{5.15} \\
\bottomrule
\end{tabular}
\end{table}

\begin{table}[t]
\centering
\caption{Per-class IoU comparison of different projection modes on SUM. The best result for each class is highlighted in bold.}
\label{tab:sum_mode_per_class_iou}
\renewcommand{\arraystretch}{1.12}
\setlength{\tabcolsep}{6pt}
\begin{tabular}{lcccccc}
\toprule
\textbf{Mode}
& \textbf{Terrain}& \textbf{High Vegetation}& \textbf{Building}
& \textbf{Water}
& \textbf{Vehicle}& \textbf{Boat} \\
\midrule
BEV-only
& 55.63
& 43.29
& 35.95
& 69.01
& \textbf{27.67}
& 26.83 \\

Oblique-only& 63.06
& 54.09
& 76.58
& \textbf{88.73}
& 11.95
& 59.10 \\

Fused
& \textbf{63.17}
& \textbf{63.10}
& \textbf{78.19}
& 85.83
& 25.75
& \textbf{61.32} \\
\bottomrule
\end{tabular}
\end{table}

\begin{table*}[t]
\centering
\caption{Per-class IoU comparison of different projection modes on H3D. The best result for each class is highlighted in bold.}
\label{tab:h3d_mode_per_class_iou}
\renewcommand{\arraystretch}{1.12}
\setlength{\tabcolsep}{4pt}
\resizebox{\textwidth}{!}{
\begin{tabular}{lccccccccccc}
\toprule
\textbf{Mode}
& \textbf{Low Veg.}
& \textbf{Imp. Surf.}
& \textbf{Vehicle}
& \textbf{Urb. Furn.}
& \textbf{Roof}
& \textbf{Facade}
& \textbf{Shrub}
& \textbf{Tree}
& \textbf{Soil/Gravel}
& \textbf{Vert. Surf.}
& \textbf{Chimney} \\
\midrule
BEV-only
& 41.82
& 47.69
& \textbf{42.51}
& 6.63
& \textbf{73.63}
& 0.48
& 4.44
& 16.34
& 29.88
& 2.30
& 0.16 \\

Oblique-only& \textbf{58.39}
& \textbf{65.15}
& 17.02
& 5.48
& 60.75
& 10.76
& 2.09
& 31.34
& \textbf{39.31}
& \textbf{8.74}
& \textbf{0.37} \\

Fused
& 54.04
& 56.55
& 35.24
& \textbf{8.00}
& 71.49
& \textbf{12.04}
& \textbf{5.01}
& \textbf{31.82}
& 29.93
& 7.77
& 0.34 \\
\bottomrule
\end{tabular}
}
\end{table*}

\begin{table*}[t]
\centering
\caption{Per-class IoU comparison of different projection modes on CUS3D. The best result for each class is highlighted in bold.}
\label{tab:cus3d_mode_per_class_iou}
\renewcommand{\arraystretch}{1.12}
\setlength{\tabcolsep}{4.5pt}
\resizebox{\textwidth}{!}{
\begin{tabular}{lcccccccccc}
\toprule
\textbf{Mode}
& \textbf{Building}& \textbf{Road}& \textbf{Car}& \textbf{Grass}& \textbf{High Veg.}& \textbf{Playground}& \textbf{Water}
& \textbf{Farmland}
& \textbf{Bldg. sites}
& \textbf{Ground} \\
\midrule
BEV-only
& 36.89
& 54.92
& 38.39
& 29.46
& 40.48
& \textbf{14.19}
& 67.25
& 31.90
& \textbf{14.39}
& 3.29 \\

Oblique-only& 46.84
& 51.97
& 22.96
& 25.14
& 39.77
& 3.37
& \textbf{79.84}
& \textbf{57.94}
& 3.35
& 3.87 \\

Fused
& \textbf{58.74}
& \textbf{57.58}
& \textbf{39.13}
& \textbf{33.71}
& \textbf{53.70}
& 6.24
& 78.06
& 53.92
& 3.68
& \textbf{8.02} \\
\bottomrule
\end{tabular}
}
\end{table*}

\subsubsection{Analysis of reliability score}
\begin{figure}[h]
\centering
\includegraphics[width=1.0\textwidth]{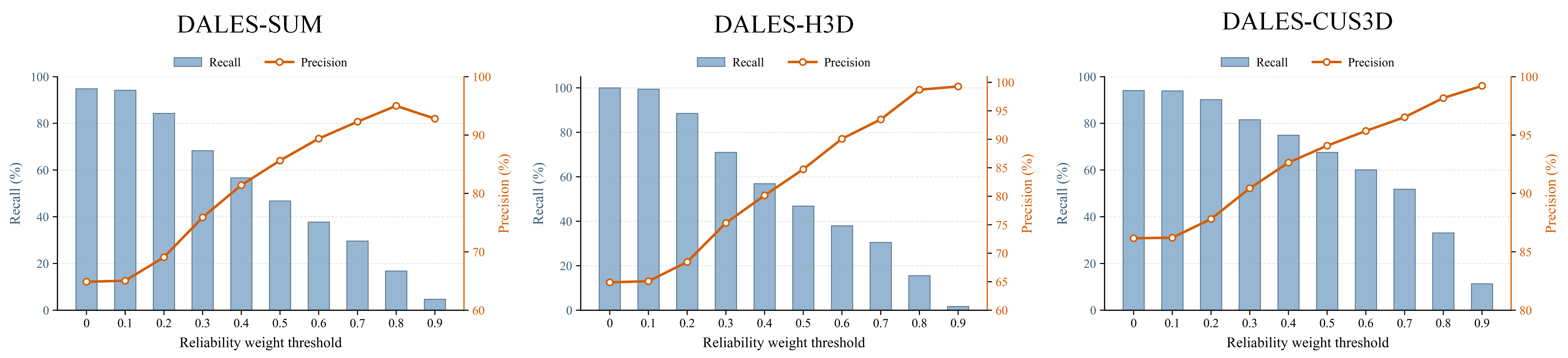}
\caption{Overall pseudo-label quality changes w.r.t. to the reliability weight thresholds across SUM, H3D, and CUS3D. Bars show pseudo-label recall, and plots show pseudo-label precision with different reliability thresholds.}\label{fig:reliability_weight}
\end{figure}

The reliability weight provides an effective criterion for VLMs pseudo-label selection. Since VLM operates under domain shifts, the reliability weight plays a crucial role in distinguishing trustworthy semantic seeds from unreliable VLM predictions. To assess whether the proposed reliability weight can accurately reflect pseudo-label quality, we analyze the precision and recall of selected pseudo labels under different reliability thresholds on SUM, H3D, and CUS3D. As shown in Fig.~\ref{fig:reliability_weight}, a clear precision--recall trade-off can be observed across all three benchmarks. On the SUM dataset, increasing the threshold consistently improves precision from $86$\% to around $99$\%, while recall decreases from $94$\% to a small fraction of points. A similar trend is observed on H3D and CUS3D. These consistent trends across datasets indicate that larger reliability weights are correlated with higher pseudo-label correctness, confirming that the proposed reliability score provides an effective quality indicator for VLM-derived semantics. 

Meanwhile, Fig.~\ref{fig:reliability_weight} also reveals the limitation of hard reliability thresholding. Although a high threshold can suppress noisy pseudo labels and improve precision, it also discards many valid target points, leading to insufficient semantic coverage. This is undesirable for cluster-centric propagation. Therefore, instead of relying only on hard selection, COSTA sets a low threshold for the cluster bank (in Eq.~(\ref{eq:cluster_bank})) followed by a soft reliability-weighted voting strategy during cluster propagation. 

\subsubsection{Ablation study}
We conduct ablation study on the DALES-SUM benchmark to validate the contribution of each component in COSTA. The \textit{Base} setting starts with hybrid projection and VLMs pseudo-label generation without TTA or cluster-centric propagation, which is based on Sa2VA~\citep{yuan2025sa2va} and hybrid multi-view projection strategy (Sec.~\ref{sec:hybrid}). As shown in Table~\ref{tab:ablation_dales_sum}, it obtains $62.89\%$ mIoU and $81.07\%$ OA, which has been demonstrated in Sec.~\ref{sec:multi-view projection}. By introducing BN-only TTA with entropy minimization and diversity regularization, followed by cluster-centric propagation, the \textit{CP} setting improves mIoU to $65.96\%$ and OA to $86.38\%$, yielding gains of $3.07\%$ and $5.31\%$, respectively. This confirms that adapting the feature distribution at inference time helps form more coherent target-domain clusters, while cluster-level propagation effectively reduces point-wise pseudo-label noise. Adding the OOD discovery loss further improves performance to $67.80\%$ mIoU and $87.42\%$ OA. It demonstrates that explicitly encouraging high uncertainty for pseudo-OOD samples prevents target-specific classes from being over-confidently absorbed into source classes, which is crucial for OSTTA. Finally, the full model equipped with reliability-weighted cluster-centric propagation achieves the best performance, reaching $70.09\%$ mIoU and $88.52\%$ OA. Compared with \textit{OOD+CP}, it further improves mIoU by $2.29\%$, demonstrating that treating all pseudo-label seeds equally is suboptimal. By weighting cluster votes according to pseudo-label reliability, COSTA suppresses unreliable VLM predictions and allows high-confidence semantic evidence to dominate the cluster decision. Overall, the full pipeline improves over the base setting by $7.20\%$ mIoU and $7.45\%$ OA, validating the contribution of each component in COSTA.

\begin{table}[t]
\centering
\caption{Ablation study of the proposed pipeline on the DALES-SUM benchmark.}
\label{tab:ablation_dales_sum}
\renewcommand{\arraystretch}{1.15}
\setlength{\tabcolsep}{7pt}
\begin{tabular}{@{}lccccc cc@{}}
\toprule
\textbf{Setting} 
& \textbf{H+S} 
& \textbf{Ent.} 
& \textbf{Div.} 
& \textbf{OOD} 
& \textbf{RW-CP} 
& \textbf{mIoU (\%)} 
& \textbf{OA (\%)} \\
\midrule
Base 
& \checkmark 
& -- 
& -- 
& -- 
& -- 
& 62.89 
& 81.07 \\

CP 
& \checkmark 
& \checkmark 
& \checkmark 
& -- 
& -- 
& 65.96 
& 86.38 \\

OOD+CP 
& \checkmark 
& \checkmark 
& \checkmark 
& \checkmark 
& -- 
& 67.80 
& 87.42 \\

Full 
& \checkmark 
& \checkmark 
& \checkmark 
& \checkmark 
& \checkmark 
& \textbf{70.09} 
& \textbf{88.52} \\
\botrule
\end{tabular}

\footnotetext{
H+S denotes hybrid projection with Sa2VA pseudo-label generation. 
Ent. and Div. denote entropy minimization and diversity regularization used in BN-only test-time adaptation. 
OOD denotes the out-of-distribution entropy. 
CP denotes cluster-centric propagation, and RW-CP denotes reliability-weighted cluster-centric propagation.
}
\end{table}

\subsubsection{Discovery of the underlying semantics}
\begin{figure}[htb]
\centering
\includegraphics[width=0.95\textwidth]{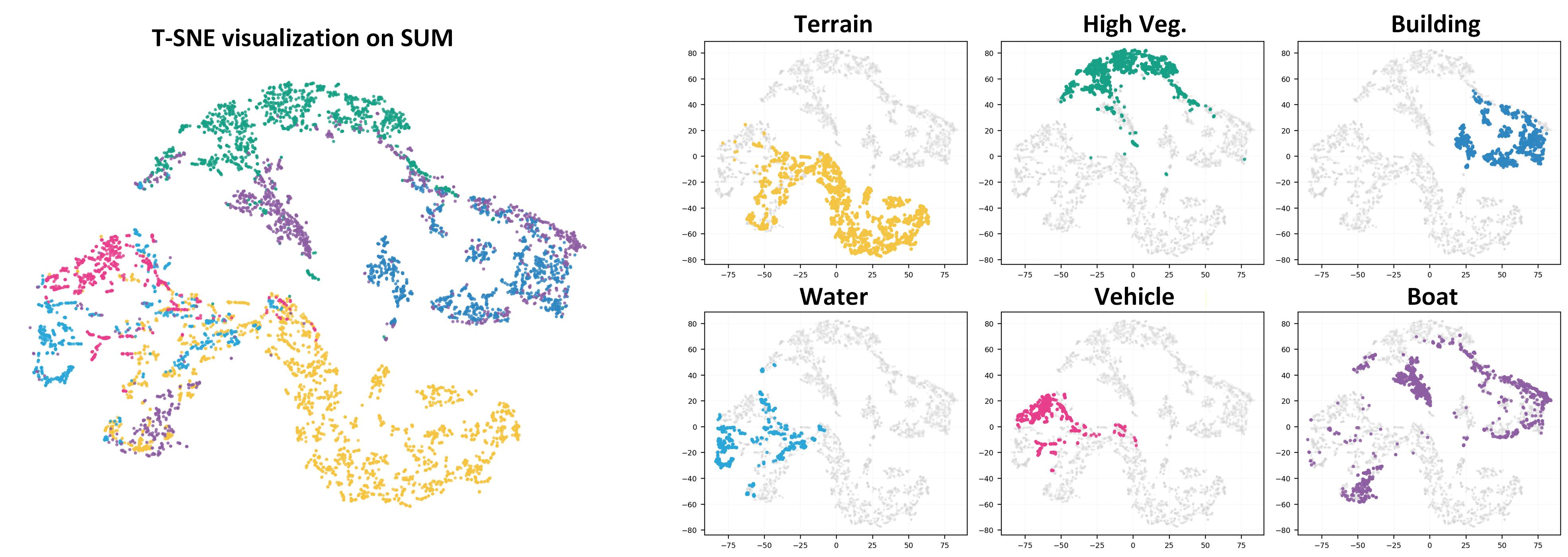}
\caption{Feature space visualization for target classes on SUM. Our method produces a well-structured feature space for target-specific classes.}\label{fig:TSNE}
\end{figure}

To assess whether our method learns to aggregate samples from target classes into different clusters, we visualize the feature representations of the SUM test samples via t-SNE~\citep{van2008visualizing} in Fig.~\ref{fig:TSNE}. It can be observed that the adapted feature space exhibits a clear and well-structured organization. Each class occupies distinct regions in the embedding space. In particular, \cls{terrain}, \cls{high vegetation}, and \cls{building} form relatively compact and separable manifolds, suggesting that the adapted model can effectively capture their semantic regularities under domain shift. Furthermore, target-specific \cls{water} and \cls{boat} also exhibit small and identifiable semantic structures, allowing them to be clustered into identifiable groups. This indicates that our method has learned the underlying semantics of known and unseen classes by producing a well-structured feature space for target-specific classes, which also supports our observation in Fig.~\ref{fig:cluster_purity_intro}.

\section{Discussion and Conclusion}\label{sec6}
We investigate aerial point cloud semantic segmentation in an open-set test-time adaptation setting (OSTTA), for which we establish a new evaluation protocol covering both domain shifts and semantic shifts. We demonstrate that COSTA achieves better performance than direct open-vocabulary transfer and competitive results against fully-supervised methods, while requiring no target annotations, source-data replay, or offline retraining. In addition to better generalization across domains, COSTA addresses the fundamental closed-set limitation of existing test-time adaptation methods by supporting user-specified target classes. Furthermore, we observe that adapted features naturally form semantically pure clusters transferable across label spaces, enabling better cross-semantics generalization.

The advantages and limitations of COSTA are inherently related. The proposed hybrid multi-view rendering allows 2D VLMs to be seamlessly applied to aerial point clouds through prompt-based segmentation without well-aligned synchronously acquired images. This design improves the flexibility of OSTTA, but also makes the quality of VLM-derived pseudo labels dependent on point density, visibility, and rendering quality. As a result, COSTA remains less effective for small-scale, sparse, or fine-grained categories, such as \cls{urban furniture}, \cls{facade}, and \cls{chimney}. These classes are more sensitive to occlusion, poor rendering quality, and ambiguous semantic boundaries. Beyond rendering quality, a deeper issue lies in semantic ambiguity: the definitions of certain aerial point cloud categories are not well-defined. For example, \cls{urban furniture} is a heterogeneous umbrella category that may include benches, poles, lights, and other visually diverse small objects, rather than a visually atomic category with consistent appearance. In this case, a simple text prompt such as ``urban furniture'' may not provide sufficiently discriminative visual grounding for VLMs. Therefore, COSTA is effective when reliable semantic seeds and coherent feature structures can be jointly established, but remains challenged by weakly observable objects and semantically ambiguous category definitions.

These findings provide guidance for future research on open-world 3D scene understanding. We envision COSTA as a step toward a more flexible and practical paradigm for aerial point cloud understanding in open environments, where perception models can adapt to new geographic regions, sensors, and task-specific category definitions without expensive re-annotation or re-training. Future work could be extended to a unified end-to-end approach for online discovery of novel classes to enable continual learning in real-world deployments. In particular, 3D Gaussian Splatting offers a promising 2D-3D intermediate representation to model geometry, appearance, opacity, and visibility in a continuous and differentiable form. Compared with direct point-to-image projection, Gaussian-based rendering may provide denser and more realistic views, better preserve occlusion relationships, and accumulate multi-view VLM evidence at the Gaussian level before propagating. We hope that our insights inspire future investigation and contribute to building models capable of generalizing to unseen environments.

\backmatter

\bmhead{Acknowledgements}

This work is supported by Jing-Jin-Ji Regional Integrated Environmental Improvement National Science and Technology Major Project (No.2026ZD1214401).

\section*{Declarations}
The authors have no relevant financial or non-financial interests to disclose.

\noindent






\bibliography{sn-bibliography}

@inproceedings{qi2017pointnet,
  title={Pointnet: Deep learning on point sets for 3d classification and segmentation},
  author={Qi, Charles R and Su, Hao and Mo, Kaichun and Guibas, Leonidas J},
  booktitle={Proceedings of the IEEE conference on computer vision and pattern recognition},
  pages={652--660},
  year={2017}
}

@article{qi2017pointnet++,
  title={Pointnet++: Deep hierarchical feature learning on point sets in a metric space},
  author={Qi, Charles Ruizhongtai and Yi, Li and Su, Hao and Guibas, Leonidas J},
  journal={Advances in neural information processing systems},
  volume={30},
  year={2017}
}

@inproceedings{landrieu2018large,
  title={Large-scale point cloud semantic segmentation with superpoint graphs},
  author={Landrieu, Loic and Simonovsky, Martin},
  booktitle={Proceedings of the IEEE conference on computer vision and pattern recognition},
  pages={4558--4567},
  year={2018}
}

@inproceedings{graham20183d,
  title={3d semantic segmentation with submanifold sparse convolutional networks},
  author={Graham, Benjamin and Engelcke, Martin and Van Der Maaten, Laurens},
  booktitle={Proceedings of the IEEE conference on computer vision and pattern recognition},
  pages={9224--9232},
  year={2018}
}

@inproceedings{hu2022sqn,
  title={Sqn: Weakly-supervised semantic segmentation of large-scale 3d point clouds},
  author={Hu, Qingyong and Yang, Bo and Fang, Guangchi and Guo, Yulan and Leonardis, Ale{\v{s}} and Trigoni, Niki and Markham, Andrew},
  booktitle={European conference on computer vision},
  pages={600--619},
  year={2022},
  organization={Springer}
}

@inproceedings{hu2020randla,
  title={Randla-net: Efficient semantic segmentation of large-scale point clouds},
  author={Hu, Qingyong and Yang, Bo and Xie, Linhai and Rosa, Stefano and Guo, Yulan and Wang, Zhihua and Trigoni, Niki and Markham, Andrew},
  booktitle={Proceedings of the IEEE/CVF conference on computer vision and pattern recognition},
  pages={11108--11117},
  year={2020}
}

@article{qian2022pointnext,
  title={Pointnext: Revisiting pointnet++ with improved training and scaling strategies},
  author={Qian, Guocheng and Li, Yuchen and Peng, Houwen and Mai, Jinjie and Hammoud, Hasan and Elhoseiny, Mohamed and Ghanem, Bernard},
  journal={Advances in neural information processing systems},
  volume={35},
  pages={23192--23204},
  year={2022}
}

@inproceedings{choy20194d,
  title={4d spatio-temporal convnets: Minkowski convolutional neural networks},
  author={Choy, Christopher and Gwak, JunYoung and Savarese, Silvio},
  booktitle={Proceedings of the IEEE/CVF conference on computer vision and pattern recognition},
  pages={3075--3084},
  year={2019}
}

@inproceedings{thomas2019kpconv,
  title={Kpconv: Flexible and deformable convolution for point clouds},
  author={Thomas, Hugues and Qi, Charles R and Deschaud, Jean-Emmanuel and Marcotegui, Beatriz and Goulette, Fran{\c{c}}ois and Guibas, Leonidas J},
  booktitle={Proceedings of the IEEE/CVF international conference on computer vision},
  pages={6411--6420},
  year={2019}
}

@inproceedings{tang2020searching,
  title={Searching efficient 3d architectures with sparse point-voxel convolution},
  author={Tang, Haotian and Liu, Zhijian and Zhao, Shengyu and Lin, Yujun and Lin, Ji and Wang, Hanrui and Han, Song},
  booktitle={European conference on computer vision},
  pages={685--702},
  year={2020},
  organization={Springer}
}

@inproceedings{zhao2021point,
  title={Point transformer},
  author={Zhao, Hengshuang and Jiang, Li and Jia, Jiaya and Torr, Philip HS and Koltun, Vladlen},
  booktitle={Proceedings of the IEEE/CVF international conference on computer vision},
  pages={16259--16268},
  year={2021}
}

@article{wu2022point,
  title={Point transformer v2: Grouped vector attention and partition-based pooling},
  author={Wu, Xiaoyang and Lao, Yixing and Jiang, Li and Liu, Xihui and Zhao, Hengshuang},
  journal={Advances in Neural Information Processing Systems},
  volume={35},
  pages={33330--33342},
  year={2022}
}

@inproceedings{wu2024point,
  title={Point transformer v3: Simpler faster stronger},
  author={Wu, Xiaoyang and Jiang, Li and Wang, Peng-Shuai and Liu, Zhijian and Liu, Xihui and Qiao, Yu and Ouyang, Wanli and He, Tong and Zhao, Hengshuang},
  booktitle={Proceedings of the IEEE/CVF conference on computer vision and pattern recognition},
  pages={4840--4851},
  year={2024}
}

@article{yang2025swin3d,
  title={Swin3d: A pretrained transformer backbone for 3d indoor scene understanding},
  author={Yang, Yu-Qi and Guo, Yu-Xiao and Xiong, Jian-Yu and Liu, Yang and Pan, Hao and Wang, Peng-Shuai and Tong, Xin and Guo, Baining},
  journal={Computational Visual Media},
  volume={11},
  number={1},
  pages={83--101},
  year={2025},
  publisher={TUP}
}

@article{wang2023octformer,
  title={Octformer: Octree-based transformers for 3d point clouds},
  author={Wang, Peng-Shuai},
  journal={ACM Transactions on Graphics (TOG)},
  volume={42},
  number={4},
  pages={1--11},
  year={2023},
  publisher={Acm New York, NY, USA}
}

@inproceedings{kolodiazhnyi2024oneformer3d,
  title={Oneformer3d: One transformer for unified point cloud segmentation},
  author={Kolodiazhnyi, Maxim and Vorontsova, Anna and Konushin, Anton and Rukhovich, Danila},
  booktitle={Proceedings of the IEEE/CVF Conference on Computer Vision and Pattern Recognition},
  pages={20943--20953},
  year={2024}
}

@inproceedings{wang2021tent,
  title={Tent: Fully Test-Time Adaptation by Entropy Minimization},
  author={Wang, Dequan and Shelhamer, Evan and Liu, Shaoteng and Olshausen, Bruno and Darrell, Trevor},
  booktitle={International Conference on Learning Representations},
  year={2021},
  url={https://openreview.net/forum?id=uXl3bZLkr3c}
}

@inproceedings{niu2022efficient,
  title={Efficient test-time model adaptation without forgetting},
  author={Niu, Shuaicheng and Wu, Jiaxiang and Zhang, Yifan and Chen, Yaofo and Zheng, Shijian and Zhao, Peilin and Tan, Mingkui},
  booktitle={International conference on machine learning},
  pages={16888--16905},
  year={2022},
  organization={PMLR}
}

@inproceedings{wang2022continual,
  title={Continual Test-Time Domain Adaptation},
  author={Wang, Qin and Fink, Olga and Van Gool, Luc and Dai, Dengxin},
  booktitle={Proceedings of Conference on Computer Vision and Pattern Recognition},
  year={2022}
}

@inproceedings{saltori2022gipso,
  title={Gipso: Geometrically informed propagation for online adaptation in 3d lidar segmentation},
  author={Saltori, Cristiano and Krivosheev, Evgeny and Lathuili{\'e}re, St{\'e}phane and Sebe, Nicu and Galasso, Fabio and Fiameni, Giuseppe and Ricci, Elisa and Poiesi, Fabio},
  booktitle={European Conference on Computer Vision},
  pages={567--585},
  year={2022},
  organization={Springer}
}

@inproceedings{zou2024hgl,
  title={Hgl: Hierarchical geometry learning for test-time adaptation in 3d point cloud segmentation},
  author={Zou, Tianpei and Qu, Sanqing and Li, Zhijun and Knoll, Alois and He, Lianghua and Chen, Guang and Jiang, Changjun},
  booktitle={European Conference on Computer Vision},
  pages={19--36},
  year={2024},
  organization={Springer}
}

@article{wang2025test,
  title={Test-time adaptation for geospatial point cloud semantic segmentation with distinct domain shifts},
  author={Wang, Puzuo and Yao, Wei and Shao, Jie and He, Zhiyi},
  journal={ISPRS Journal of Photogrammetry and Remote Sensing},
  volume={229},
  pages={422--435},
  year={2025},
  publisher={Elsevier}
}

@inproceedings{weijler2025ttt,
  title={Ttt-kd: Test-time training for 3d semantic segmentation through knowledge distillation from foundation models},
  author={Weijler, Lisa and Mirza, M Jehanzeb and Sick, Leon and Ekkazan, Can and Hermosilla, Pedro},
  booktitle={2025 International Conference on 3D Vision (3DV)},
  pages={1264--1274},
  year={2025},
  organization={IEEE}
}

@article{gong2022note,
  title={Note: Robust continual test-time adaptation against temporal correlation},
  author={Gong, Taesik and Jeong, Jongheon and Kim, Taewon and Kim, Yewon and Shin, Jinwoo and Lee, Sung-Ju},
  journal={Advances in Neural Information Processing Systems},
  volume={35},
  pages={27253--27266},
  year={2022}
}

@inproceedings{yuan2023robust,
  title={Robust test-time adaptation in dynamic scenarios},
  author={Yuan, Longhui and Xie, Binhui and Li, Shuang},
  booktitle={Proceedings of the IEEE/CVF Conference on Computer Vision and Pattern Recognition},
  pages={15922--15932},
  year={2023}
}

@article{niu2023towards,
  title={Towards stable test-time adaptation in dynamic wild world},
  author={Niu, Shuaicheng and Wu, Jiaxiang and Zhang, Yifan and Wen, Zhiquan and Chen, Yaofo and Zhao, Peilin and Tan, Mingkui},
  journal={arXiv preprint arXiv:2302.12400},
  year={2023}
}

@inproceedings{li2023robustness,
  title={On the robustness of open-world test-time training: Self-training with dynamic prototype expansion},
  author={Li, Yushu and Xu, Xun and Su, Yongyi and Jia, Kui},
  booktitle={Proceedings of the IEEE/CVF International Conference on Computer Vision},
  pages={11836--11846},
  year={2023}
}

@inproceedings{gao2024unified,
  title={Unified entropy optimization for open-set test-time adaptation},
  author={Gao, Zhengqing and Zhang, Xu-Yao and Liu, Cheng-Lin},
  booktitle={Proceedings of the IEEE/CVF Conference on Computer Vision and Pattern Recognition},
  pages={23975--23984},
  year={2024}
}

@article{dong2025towards,
  title={Towards robust multimodal open-set test-time adaptation via adaptive entropy-aware optimization},
  author={Dong, Hao and Chatzi, Eleni and Fink, Olga},
  journal={arXiv preprint arXiv:2501.13924},
  year={2025}
}

@article{lee2025stabilizing,
  title={Stabilizing Open-Set Test-Time Adaptation via Primary-Auxiliary Filtering and Knowledge-Integrated Prediction},
  author={Lee, Byung-Joon and Lee, Jin-Seop and Lee, Jee-Hyong},
  journal={arXiv preprint arXiv:2508.18751},
  year={2025}
}

@inproceedings{zou2026good,
  title={GOOD: Geometry-guided out-of-distribution modeling for open-set test-time adaptation in point cloud semantic segmentation},
  author={Zou, Tianpei and Yu, Guo and Wu, Ya and Lu, Fan and Xu, Zhongcong and Bo, Zhang and Wang, Ziqiao and Qu, Sanqing and Chen, Guang},
  booktitle={Proc. Int. Conf. Learn. Representations},
  year={2026}
}

@inproceedings{varney2020dales,
  title={DALES: A large-scale aerial LiDAR data set for semantic segmentation},
  author={Varney, Nina and Asari, Vijayan K and Graehling, Quinn},
  booktitle={Proceedings of the IEEE/CVF conference on computer vision and pattern recognition workshops},
  pages={186--187},
  year={2020}
}

@article{kolle2021hessigheim,
  title={The Hessigheim 3D (H3D) benchmark on semantic segmentation of high-resolution 3D point clouds and textured meshes from UAV LiDAR and Multi-View-Stereo},
  author={K{\"o}lle, Michael and Laupheimer, Dominik and Schmohl, Stefan and Haala, Norbert and Rottensteiner, Franz and Wegner, Jan Dirk and Ledoux, Hugo},
  journal={ISPRS Open Journal of Photogrammetry and Remote Sensing},
  volume={1},
  pages={100001},
  year={2021},
  publisher={Elsevier}
}

@article{sum2021,
author = {Weixiao Gao and Liangliang Nan and Bas Boom and Hugo Ledoux},
title = {SUM: A Benchmark Dataset of Semantic Urban Meshes},
journal = {ISPRS Journal of Photogrammetry and Remote Sensing},
volume = {179},
pages = {108-120},
year={2021},
issn = {0924-2716},
doi = {10.1016/j.isprsjprs.2021.07.008},
url = {https://www.sciencedirect.com/science/article/pii/S0924271621001854},
}

@article{gao2024cus3d,
  title={CUS3D: A new comprehensive urban-scale semantic-segmentation 3D benchmark dataset},
  author={Gao, Lin and Liu, Yu and Chen, Xi and Liu, Yuxiang and Yan, Shen and Zhang, Maojun},
  journal={Remote Sensing},
  volume={16},
  number={6},
  pages={1079},
  year={2024},
  publisher={MDPI}
}

@inproceedings{10.1007/978-3-031-20059-5_31,
author = {Ghiasi, Golnaz and Gu, Xiuye and Cui, Yin and Lin, Tsung-Yi},
title = {Scaling Open-Vocabulary Image Segmentation with\&nbsp;Image-Level Labels},
year = {2022},
isbn = {978-3-031-20058-8},
publisher = {Springer-Verlag},
address = {Berlin, Heidelberg},
doi = {10.1007/978-3-031-20059-5_31},
booktitle = {Computer Vision – ECCV 2022: 17th European Conference, Tel Aviv, Israel, October 23–27, 2022, Proceedings, Part XXXVI},
pages = {540–557},
numpages = {18},
location = {Tel Aviv, Israel}
}

@INPROCEEDINGS{10205470,
  author={Dong, Xiaoyi and Bao, Jianmin and Zheng, Yinglin and Zhang, Ting and Chen, Dongdong and Yang, Hao and Zeng, Ming and Zhang, Weiming and Yuan, Lu and Chen, Dong and Wen, Fang and Yu, Nenghai},
  booktitle={2023 IEEE/CVF Conference on Computer Vision and Pattern Recognition (CVPR)}, 
  title={MaskCLIP: Masked Self-Distillation Advances Contrastive Language-Image Pretraining}, 
  year={2023},
  volume={},
  number={},
  pages={10995-11005},
  doi={10.1109/CVPR52729.2023.01058}}

@inproceedings{xu2023open,
  title={Open-vocabulary panoptic segmentation with text-to-image diffusion models},
  author={Xu, Jiarui and Liu, Sifei and Vahdat, Arash and Byeon, Wonmin and Wang, Xiaolong and De Mello, Shalini},
  booktitle={Proceedings of the IEEE/CVF conference on computer vision and pattern recognition},
  pages={2955--2966},
  year={2023}
}

@inproceedings{cho2024cat,
  title={Cat-seg: Cost aggregation for open-vocabulary semantic segmentation},
  author={Cho, Seokju and Shin, Heeseong and Hong, Sunghwan and Arnab, Anurag and Seo, Paul Hongsuck and Kim, Seungryong},
  booktitle={Proceedings of the IEEE/CVF Conference on Computer Vision and Pattern Recognition},
  pages={4113--4123},
  year={2024}
}

@inproceedings{peng2023openscene,
  title={Openscene: 3d scene understanding with open vocabularies},
  author={Peng, Songyou and Genova, Kyle and Jiang, Chiyu and Tagliasacchi, Andrea and Pollefeys, Marc and Funkhouser, Thomas and others},
  booktitle={Proceedings of the IEEE/CVF conference on computer vision and pattern recognition},
  pages={815--824},
  year={2023}
}

@inproceedings{liu2023segment,
  title = {Segment Any Point Cloud Sequences by Distilling Vision Foundation Models},
  author = {Liu, Youquan and Kong, Lingdong and Cen, Jun and Chen, Runnan and Zhang, Wenwei and Pan, Liang and Chen, Kai and Liu, Ziwei},
  booktitle = {Advances in Neural Information Processing Systems}, 
  year = {2023},
}

@inproceedings{jiang2024open,
  title={Open-vocabulary 3d semantic segmentation with foundation models},
  author={Jiang, Li and Shi, Shaoshuai and Schiele, Bernt},
  booktitle={Proceedings of the IEEE/CVF Conference on Computer Vision and Pattern Recognition},
  pages={21284--21294},
  year={2024}
}

@inproceedings{yang2024regionplc,
  title={Regionplc: Regional point-language contrastive learning for open-world 3d scene understanding},
  author={Yang, Jihan and Ding, Runyu and Deng, Weipeng and Wang, Zhe and Qi, Xiaojuan},
  booktitle={Proceedings of the IEEE/CVF conference on computer vision and pattern recognition},
  pages={19823--19832},
  year={2024}
}

@inproceedings{wu2024towards,
  title={Towards large-scale 3d representation learning with multi-dataset point prompt training},
  author={Wu, Xiaoyang and Tian, Zhuotao and Wen, Xin and Peng, Bohao and Liu, Xihui and Yu, Kaicheng and Zhao, Hengshuang},
  booktitle={Proceedings of the IEEE/CVF conference on computer vision and pattern recognition},
  pages={19551--19562},
  year={2024}
}

@article{takmaz2023openmask3d,
  title={Openmask3d: Open-vocabulary 3d instance segmentation},
  author={Takmaz, Ay{\c{c}}a and Fedele, Elisabetta and Sumner, Robert W and Pollefeys, Marc and Tombari, Federico and Engelmann, Francis},
  journal={arXiv preprint arXiv:2306.13631},
  year={2023}
}

@article{wang2026openurban3d,
  title={OpenUrban3D: Label-Free Open-Vocabulary Semantic Segmentation of Large-Scale Urban Point Clouds},
  author={Wang, Chongyu and Jing, Kunlei and Zhu, Jihua and Wang, Di},
  journal={IEEE Transactions on Geoscience and Remote Sensing},
  year={2026},
  publisher={IEEE}
}

@article{xu2025cityseg,
  title={CitySeg: A 3D Open Vocabulary Semantic Segmentation Foundation Model in City-scale Scenarios},
  author={Xu, Jialei and Wei, Zizhuang and You, Weikang and Li, Linyun and Sun, Weijian},
  journal={arXiv preprint arXiv:2508.09470},
  year={2025}
}

@article{gao2022piie,
  title={PIIE-DSA-Net for 3D semantic segmentation of urban indoor and outdoor datasets},
  author={Gao, Fengjiao and Yan, Yiming and Lin, Hemin and Shi, Ruiyao},
  journal={Remote Sensing},
  volume={14},
  number={15},
  pages={3583},
  year={2022},
  publisher={MDPI}
}

@inproceedings{ding2023pla,
  title={Pla: Language-driven open-vocabulary 3d scene understanding},
  author={Ding, Runyu and Yang, Jihan and Xue, Chuhui and Zhang, Wenqing and Bai, Song and Qi, Xiaojuan},
  booktitle={Proceedings of the IEEE/CVF conference on computer vision and pattern recognition},
  pages={7010--7019},
  year={2023}
}

@article{gao2026source,
  title={Source-Free Domain Adaptation for Geospatial Point Cloud Semantic Segmentation},
  author={Gao, Yuan and Cao, Di and Xi, Xiaohuan and Nie, Sheng and Xia, Shaobo and Wang, Cheng},
  journal={arXiv preprint arXiv:2601.08375},
  year={2026}
}

@inproceedings{wu2025sonata,
  title={Sonata: Self-supervised learning of reliable point representations},
  author={Wu, Xiaoyang and DeTone, Daniel and Frost, Duncan and Shen, Tianwei and Xie, Chris and Yang, Nan and Engel, Jakob and Newcombe, Richard and Zhao, Hengshuang and Straub, Julian},
  booktitle={Proceedings of the Computer Vision and Pattern Recognition Conference},
  pages={22193--22204},
  year={2025}
}

@misc{carion2025sam3segmentconcepts,
      title={SAM 3: Segment Anything with Concepts},
      author={Nicolas Carion and Laura Gustafson and Yuan-Ting Hu and Shoubhik Debnath and Ronghang Hu and Didac Suris and Chaitanya Ryali and Kalyan Vasudev Alwala and Haitham Khedr and Andrew Huang and Jie Lei and Tengyu Ma and Baishan Guo and Arpit Kalla and Markus Marks and Joseph Greer and Meng Wang and Peize Sun and Roman Rädle and Triantafyllos Afouras and Effrosyni Mavroudi and Katherine Xu and Tsung-Han Wu and Yu Zhou and Liliane Momeni and Rishi Hazra and Shuangrui Ding and Sagar Vaze and Francois Porcher and Feng Li and Siyuan Li and Aishwarya Kamath and Ho Kei Cheng and Piotr Dollár and Nikhila Ravi and Kate Saenko and Pengchuan Zhang and Christoph Feichtenhofer},
      year={2025},
      eprint={2511.16719},
      archivePrefix={arXiv},
      primaryClass={cs.CV},
      url={https://arxiv.org/abs/2511.16719},
}

@article{yuan2025sa2va,
  title={Sa2va: Marrying sam2 with llava for dense grounded understanding of images and videos},
  author={Yuan, Haobo and Li, Xiangtai and Zhang, Tao and Sun, Yueyi and Huang, Zilong and Xu, Shilin and Ji, Shunping and Tong, Yunhai and Qi, Lu and Feng, Jiashi and others},
  journal={arXiv preprint arXiv:2501.04001},
  year={2025}
}

@article{alami2026open,
  title={Open-Vocabulary Segmentation of Aerial Point Clouds},
  author={Alami, Ashkan and Remondino, Fabio},
  journal={Remote Sensing},
  volume={18},
  number={4},
  pages={572},
  year={2026},
  publisher={MDPI}
}

@article{lin2026towards,
  title={Towards Open-Vocabulary ALS Point Clouds Semantic Segmentation: An Empirical Study},
  author={Lin, Yanghong and Li, Tianyu and Zhou, Shudong and Zhang, Jingru and Fang, Li and Yao, Wei},
  journal={The International Archives of the Photogrammetry, Remote Sensing and Spatial Information Sciences},
  volume={49},
  pages={223--229},
  year={2026},
  publisher={Copernicus GmbH}
}

@article{wang2022new,
  title={A new weakly supervised approach for ALS point cloud semantic segmentation},
  author={Wang, Puzuo and Yao, Wei},
  journal={ISPRS Journal of Photogrammetry and Remote Sensing},
  volume={188},
  pages={237--254},
  year={2022},
  publisher={Elsevier}
}

@article{shao2024urban,
  title={Urban GeoBIM construction by integrating semantic LiDAR point clouds with as-designed BIM models},
  author={Shao, Jie and Yao, Wei and Wang, Puzuo and He, Zhiyi and Luo, Lei},
  journal={IEEE Transactions on Geoscience and Remote Sensing},
  volume={62},
  pages={1--12},
  year={2024},
  publisher={IEEE}
}

@inproceedings{gao2026openfly,
  title={Openfly: A comprehensive platform for aerial vision-language navigation},
  author={Gao, Yunpeng and Li, Chenhui and You, Zhongrui and Liu, Junli and Zhen, Li and Chen, Pengan and Chen, Qizhi and Tang, Zhonghan and Wang, Liansheng and Tang, Yiwen and others},
  booktitle={International Conference on Learning Representations},
  volume={2026},
  pages={101858--101871},
  year={2026}
}

@article{van2008visualizing,
  title={Visualizing data using t-SNE.},
  author={Van der Maaten, Laurens and Hinton, Geoffrey},
  journal={Journal of machine learning research},
  volume={9},
  number={11},
  year={2008}
}

@article{hou2026agc,
  title={Agc-drive: A large-scale dataset for real-world aerial-ground collaboration in driving scenarios},
  author={Hou, Yunhao and Zou, Bochao and Zhang, Min and Yang, Shangdong and Zhang, Yanmei and Zhuo, Junbao and Chen, Siheng and Chen, Jiansheng and Ma, Huimin},
  journal={Advances in Neural Information Processing Systems},
  volume={38},
  year={2026}
}

@article{hu2022sensaturban,
  title={Sensaturban: Learning semantics from urban-scale photogrammetric point clouds},
  author={Hu, Qingyong and Yang, Bo and Khalid, Sheikh and Xiao, Wen and Trigoni, Niki and Markham, Andrew},
  journal={International Journal of Computer Vision},
  volume={130},
  number={2},
  pages={316--343},
  year={2022},
  publisher={Springer}
}

@article{POLEWSKI2015252,
title = {Detection of fallen trees in ALS point clouds using a Normalized Cut approach trained by simulation},
journal = {ISPRS Journal of Photogrammetry and Remote Sensing},
volume = {105},
pages = {252-271},
year = {2015},
issn = {0924-2716},
doi = {https://doi.org/10.1016/j.isprsjprs.2015.01.010},
url = {https://www.sciencedirect.com/science/article/pii/S0924271615000271},
author = {Przemyslaw Polewski and Wei Yao and Marco Heurich and Peter Krzystek and Uwe Stilla}
}

@article{YAO2011260,
title = {Extraction and motion estimation of vehicles in single-pass airborne LiDAR data towards urban traffic analysis},
journal = {ISPRS Journal of Photogrammetry and Remote Sensing},
volume = {66},
number = {3},
pages = {260-271},
year = {2011},
issn = {0924-2716},
doi = {https://doi.org/10.1016/j.isprsjprs.2010.10.005},
url = {https://www.sciencedirect.com/science/article/pii/S0924271610000961},
author = {Wei Yao and Stefan Hinz and Uwe Stilla}
}

@article{xiang2024efficient,
  title={Efficient high-quality vectorized modeling of large-scale scenes},
  author={Xiang, Xiaojun and Jiang, Hanqing and Yu, Yihao and Shen, Donghui and Zhen, Jianan and Bao, Hujun and Zhou, Xiaowei and Zhang, Guofeng},
  journal={International Journal of Computer Vision},
  volume={132},
  number={10},
  pages={4564--4588},
  year={2024},
  publisher={Springer}
}

@article{li2025deep,
  title={Deep hierarchical learning for 3d semantic segmentation},
  author={Li, Chongshou and Liu, Yuheng and Li, Xinke and Zhang, Yuning and Li, Tianrui and Yuan, Junsong},
  journal={International Journal of Computer Vision},
  volume={133},
  number={7},
  pages={4420--4441},
  year={2025},
  publisher={Springer}
}

@ARTICLE{10008011,
  author={Zhu, Yilin and Kong, Yang and Jie, Yingrui and Xu, Shiyou and Cheng, Hui},
  journal={IEEE Robotics and Automation Letters}, 
  title={GRACO: A Multimodal Dataset for Ground and Aerial Cooperative Localization and Mapping}, 
  year={2023},
  volume={8},
  number={2},
  pages={966-973},
  doi={10.1109/LRA.2023.3234802}}

@inproceedings{wang2025towards,
  title={Towards realistic uav vision-language navigation: Platform, benchmark, and methodology},
  author={Wang, Xiangyu and Yang, Donglin and Kwan, Hohin and Chen, Jinyu and Li, Hongsheng and Liao, Yue and Liu, Si and others},
  booktitle={International Conference on Learning Representations},
  volume={2025},
  pages={7292--7310},
  year={2025}
}

\end{document}